\documentclass[twoside,11pt]{fairmeta} 
\usepackage[inkscapelatex=false]{svg}
\usepackage{graphicx}
\usepackage{subcaption}
\usepackage{xspace}
\usepackage{enumitem}
\usepackage[T1]{fontenc}
\usepackage{wrapfig}

\graphicspath{{figures/}}

\newcommand{\tie}{\textsc{TIE}\xspace}
\newcommand{\plm}{\textsc{PLM}\xspace}
\newcommand{\ours}{\plm-\tie}

\newcommand{\bslneC}{\plm-Cont\xspace}
\newcommand{\pef}{\textsc{Perception Encoder}\xspace}
\newcommand{\pes}{\textsc{PE}\xspace}
\newcommand{\tfl}{\textsc{T5-Large}\xspace}

\newcommand{\imageencoder}{\textsc{IE}\xspace}
\newcommand{\textencoder}{\textsc{TE}\xspace}
\newcommand{\llm}{\textsc{LLM}\xspace}

\title{Text-Guided Semantic Image Encoder}
\author[1,\star]{Raghuveer Thirukovalluru}
\author[2]{Xiaochuang Han}
\author[1]{Bhuwan Dhingra}
\author[2]{Emily Dinan}
\author[2]{Maha Elbayad}
\affiliation[1]{Duke University}
\affiliation[2]{FAIR at Meta}
\contribution[\star]{Work done during an internship at Meta FAIR}
\date{November 23, 2025}
\correspondence{Maha Elbayad at \email{elbayadm@meta.com}}

\newif\ifdraft
\draftfalse

\ifdraft
    \newcommand{\rt}[1]{\textcolor{blue}{\textbf{raghu:} #1}}
    \newcommand{\emily}[1]{\textcolor{magenta}{\textbf{emily:} #1}}
    \newcommand{\maha}[1]{\textcolor{red}{\textbf{Maha:} #1}}
    \newcommand{\bd}[1]{\textcolor{cyan}{\textbf{Bhuwan:} #1}}
\else
    \newcommand{\rt}[1]{}
    \newcommand{\emily}[1]{}
    \newcommand{\maha}[1]{}
    \newcommand{\bd}[1]{}
\fi 

\begin{document}
\abstract{
\small
Image encoders, a fundamental component of vision–language models (VLMs), are typically pretrained independently before being aligned with a language model.
This standard paradigm results in encoders that process images agnostically, without regard to the specific downstream task or text query.
To address this limitation, we propose the Text-Guided Semantic Image Encoder (\tie), which generates image representations conditioned on the input text query.
VLMs equipped with \tie outperform their conventional counterparts by $+1.5$ and $+1.3$ points on average across nine image-to-text benchmarks at the $1$B and $3$B scales, respectively, with gains reaching up to $6$ points on tasks such as DocVQA and InfoVQA.
Moreover, \tie-based VLMs attain superior performance while utilizing only half as many image tiles (tokens), resulting in notably improved inference efficiency.
\tie also generalizes well with generic queries, 
indicating that text-conditioned training effectively optimizes the encoder to capture key visual features.
Qualitative analysis confirms that \tie consistently attends to query-relevant regions, enhancing both interpretability and query-specific grounding.
}

\maketitle
  
\begin{figure}[!ht]
    \centering
    \begin{subfigure}[b]{0.34\linewidth}
        \centering
        \includegraphics[width=\linewidth]{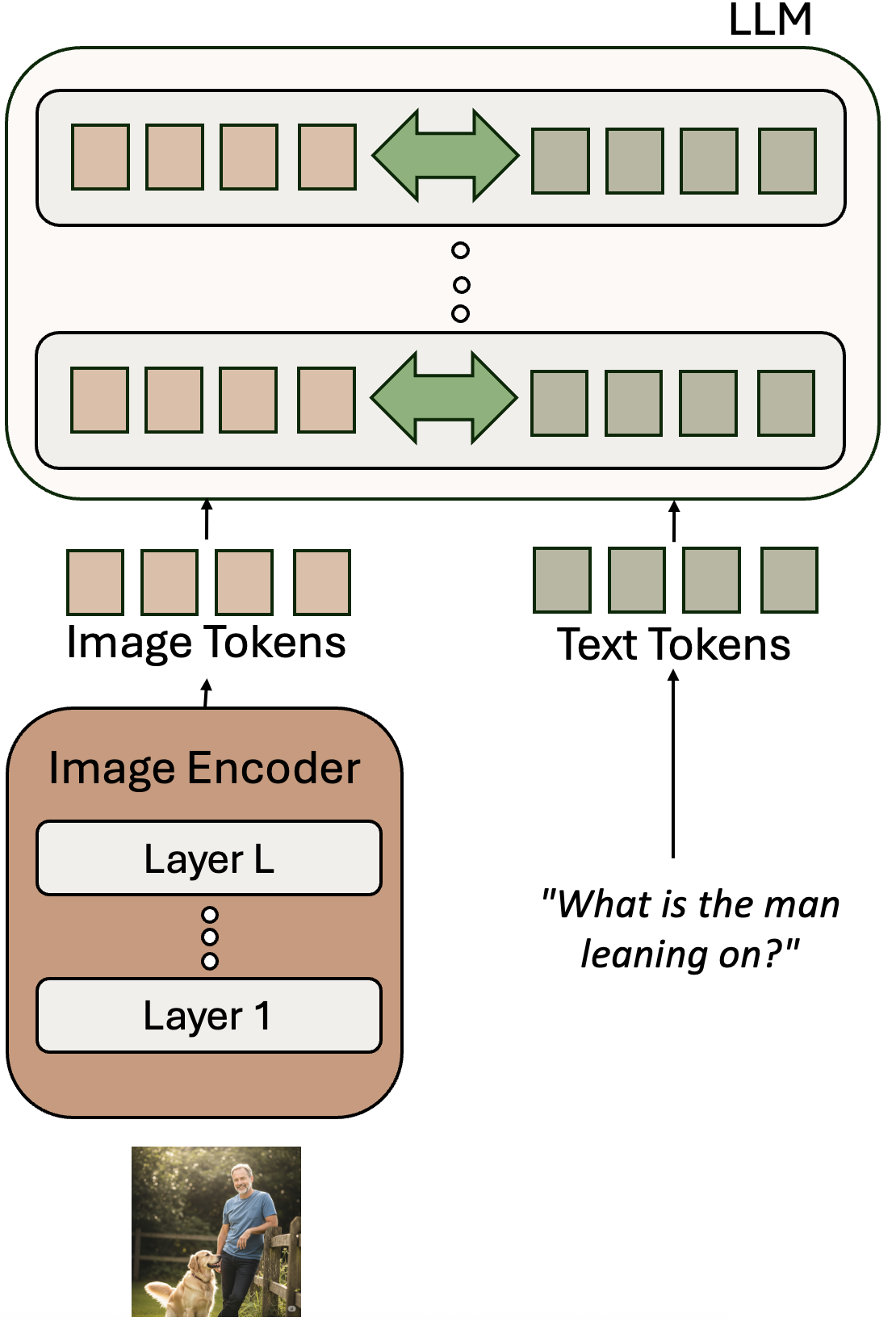}
        \caption{VLM in Prior Works}
        \label{fig:sub_a}
    \end{subfigure}
    \hspace{5pt}
    \begin{subfigure}[b]{0.34\linewidth}
        \centering
        \includegraphics[width=\linewidth]{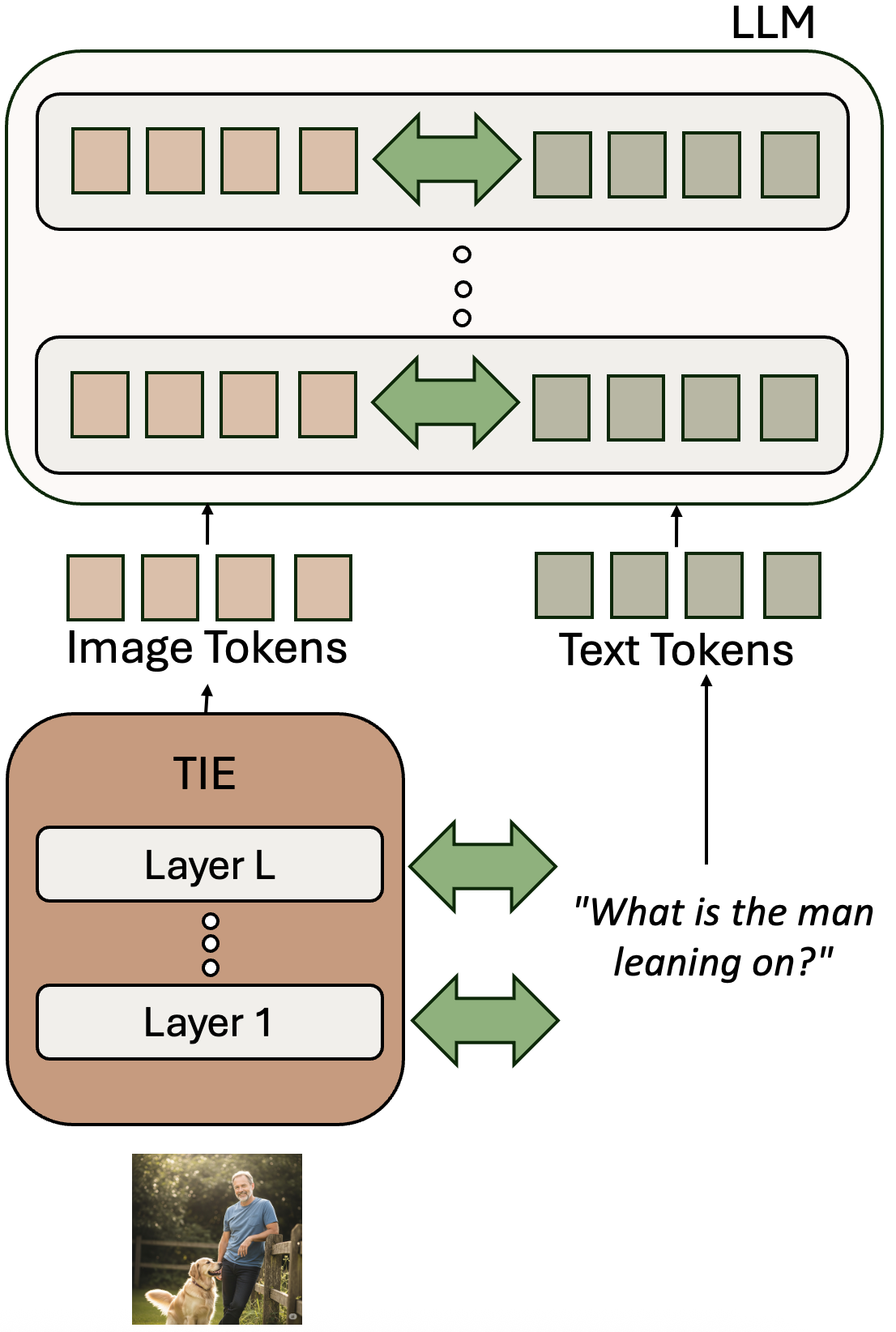}
        \caption{VLM with TIE}
        \label{fig:sub_b}
    \end{subfigure}
    \caption{VLM in (a) prior works restrict text–image interaction to the LLM layers; (b) proposed \tie image encoder, which generates image representations/tokens conditioned on the given query.}
    \label{fig:combined}
\end{figure}

\section{Introduction}

Vision–language models (VLMs) have emerged as powerful tools for addressing a wide range of multimodal tasks, such as image understanding, captioning, OCR, and visual question answering. 
A VLM typically consists of two key components: (i) an image encoder (IE), which transforms raw pixel-level information into a sequence of visual tokens, and (ii) a large language model (LLM), which processes these visual tokens alongside textual inputs to perform downstream reasoning and understanding tasks.

Image encoders are typically pretrained through large-scale contrastive learning on billions of image–text pairs~\citep{radford2021learning, zhai2023sigmoid}. In this setup, an image backbone (CNN or Transformer) encodes each image, while a text network maps its caption into a shared embedding space. Models such as \textsc{CLIP}~\citep{radford2021learning} and \textsc{SigLIP}~\citep{zhai2023sigmoid} align these representations to learn joint multimodal features. In the next stage, the text tower is removed, and the image tower serves as the IE. The IE is then aligned with a pretrained text-only LLM through \textit{language alignment}, which jointly fine-tunes both components on diverse image-to-text tasks while progressively unfreezing the IE and LLM across different stages to obtain the final VLM.

State-of-the-art VLMs---including \textsc{PerceptionLM} (\plm) ~\citep{cho2025perceptionlm}, \textsc{InternVL3}~\citep{zhu2025internvl3}, \textsc{InternVL2.5}~\citep{chen2024expanding}, and \textsc{Qwen-2.5-VL}~\citep{bai2025qwen2}---consistently follow this paradigm, establishing it as the prevailing framework for modern VLM training. Each of these models introduces distinct strategies for improving the IE: \plm demonstrates that the intermediate layers of its image encoder, rather than the final layer, yield better image representations. \textsc{InternVL2.5} and \textsc{InternVL3} leverage knowledge distillation from a larger 6B IE, to construct their smaller IE, \textsc{InternVIT-300M}.

While prior research has emphasized improving training methodologies and data quality, the underlying architectural design of image encoders has remained relatively static. In particular, likely following from the common practice of independently pretraining image encoders, most pipelines still adopt a design in which the image encoder produces general-purpose, query-agnostic image features that are consumed by the VLM. Such query-agnostic representations are not ideal. For example, consider an image with a dog, a man leaning on a fence, and bushes on either side. To answer a question like “\emph{What is the man leaning on?},” the image encoder should focus on the region surrounding the man rather than irrelevant areas such as the bushes or the sky. However, current image encoders remain query-agnostic, treating all regions of an image equally with the same static processing.
Consequently, they often rely on a large number of tiles (each tile encoded with a fixed number of tokens, e.g., 256) to represent an image, which is computationally expensive for multimodal reasoning and question-answering tasks.

To address this limitation, we propose the \textbf{T}ext-Guided Semantic \textbf{I}mage \textbf{E}ncoder (\tie).
\tie is an image encoder that conditions its representations on the input query (see \Cref{fig:combined}).
It uses
query/task embeddings derived from an external language encoder (e.g., \tfl) or the VLM’s own input embeddings,
to produce semantically enriched visual features aligned with the target image–text objective.
Our key findings and contributions are as follows:
\begin{itemize}
\item[--] VLMs integrated with \tie, which we call \ours, achieve average improvements of $+1.5$ and $+1.3$ points compared to the baseline, \bslneC~\citep{cho2025perceptionlm} (\plm continually finetuned with exactly identical conditions), on nine image-to-text benchmarks at $1B$ and $3B$ scales respectively. Gains reach up to $+6$ points on tasks that typically require a large number of tiles (e.g., DocVQA and InfoVQA). 
\item[--] \ours surpasses the baseline \bslneC even when operating with half the number of tiles (i.e., half the visual tokens), resulting in notably faster inference. 
\item[--] VLM's own non-contextual input text embeddings (as opposed to separate strong text encoder) are sufficient to surpass the \bslneC by $+0.9$ on the nine tasks, underscoring the potential for more unified future architectures. 
\item[--] \ours retains strong performance even with a generic query at inference, indicating that text-conditioned training yields a more effectively optimized image encoder.
\item[--] Finally, we present qualitative analyses that demonstrate \tie’s improvements over a standard IE and reveal how its attention patterns adapt appropriately to different queries.
\end{itemize}

\section{Related Work}

Image encoders are typically pretrained using contrastive learning on large-scale image-text data (in the billions), to align the representations of images and text. \textsc{CLIP}~\citep{radford2021learning} used a dual encoder setup with such contrastive data.
\textsc{SigLIP}~\citep{zhai2023sigmoid} replaced \textsc{CLIP}’s softmax-based contrastive loss with a sigmoid loss, enabling more stable and better alignment across varying batch sizes. \textsc{CLIP} and \textsc{SigLIP} were subsequently adopted in a range of VLMs
 (\textsc{Qwen-VL} \citep{bai2023qwenvlversatilevisionlanguagemodel}, \textsc{PaliGemma} \citep{beyer2024paligemma}, \textsc{BLIP2} \citep{li2023blip}, \textsc{LLAVA} \citep{liu2023visual}, \textsc{MM1} \citep{mckinzie2024mm1}, etc.) 
 where they were further aligned alongside an LLM on image-to-text tasks.

Early approaches in this line of work kept both the image encoder and the language model frozen, training only lightweight bridging modules between them.
\textsc{Flamingo}~\citep{alayrac2022flamingo} addressed variable image resolutions using a perceiver resampler that maps a variable number of image embeddings to a fixed set of tokens for the LLM.
\textsc{BLIP-2}~\citep{li2023blip} similarly used a query transformer (Q-former) with learnable query embeddings to attend to variable-length image features and produce a fixed-size representation.
\textsc{InstructBLIP}~\citep{dai2023instructblip}, closer to the proposed \tie, incorporated textual instructions into its Q-former to create query-specific image embeddings. However, text conditioning in \textsc{InstructBLIP} is limited to the bridge module between the image encoder and LLM, whereas \tie\ integrates text guidance throughout all IE layers.
Additionally, these models rely on a fixed number of learnable query embeddings ($32$ in \textsc{BLIP-2}/\textsc{InstructBLIP} and $64$ in \textsc{Flamingo}) while \tie\ supports a variable number of image tiles, enabling more flexible visual representations.

These fixed-size approaches were surpassed by methods that utilize multiple tiles per image (\textsc{InternVL2} \citep{chen2024expanding}, \textsc{InternVL3} \citep{zhu2025internvl3}, \plm~\citep{cho2025perceptionlm}), allowing the LLM to process a variable number of image tokens depending on resolution and aspect ratio. These methods typically employ a dynamic-resolution strategy in which each image is divided into multiple fixed-size tiles (e.g., $448\times448$) to preserve both the aspect ratio and the effective area of the original image. Each tile is independently processed by the image encoder. The representations from all tiles are then fed into the LLM, and the entire system is aligned on image–text tasks through language alignment. These works also highlighted the need for improved training methodologies for image encoders. 

\textsc{InternViT}, IE of \textsc{InternVL}~\citep{chen2024expanding} series of models, first contrastively pretrains a large $6$B image encoder and then distills its abilities into a smaller $300$M IE. \textsc{SigLIP2}~\citep{tschannen2025siglip} combines a sigmoid image-text alignment loss with auxiliary captioning, self-distillation, masked prediction, and data curation. 
\pef (\pes)~\citep{bolya2025perception}, image encoder of \plm series of models, demonstrated that contrastive pretraining alone is sufficient to produce state-of-the-art image encoders. It further
showed that intermediate layers of contrastively trained IEs provide more effective representations for vision–language models than their final layers. 

While prior methods rely on contrastive alignment, \textsc{AIMv2}~\citep{fini2025multimodal} introduced a multimodal decoder that generates both text tokens and image patches using reconstruction and language alignment losses instead.
More recently, \textsc{OpenVision2}~\citep{liu2025openvision} demonstrated that contrastive pretraining can be entirely removed, with language alignment alone sufficing to match contrastively trained image encoders.

Building on these advancements, we address a key limitation of existing image encoders i.e. their query-independent processing. \tie leverages the query to produce text-guided image representations, resulting in semantically richer and contextually aligned features for multimodal understanding.

\section{Background} \label{sec:background}

In this section, we provide background on the typical operation of a vision encoder within a vision--language model (VLM). Given an image of arbitrary resolution and aspect ratio, and an associated instruction/query $q$, the VLM is tasked with generating an answer $a$ that addresses the query. 
The original image is first split into multiple tiles $T$, each of a fixed size $H \times W$ (e.g., $448 \times 448$). Each tile is then independently processed by an image encoder, denoted as $\imageencoder$, which transforms the raw pixel inputs into a sequence of patches, each represented by a $d_{\imageencoder}$-dimensional embedding. With a patch resolution of $(P, P)$, the image encoder embeds the input tile $I$ into non-overlapping $N_\imageencoder{=}HW/P^2$ patches.

Note that patches within a tile can attend to each other whilst patches across tiles cannot attend to each other as they are processed independently. The $d_{\imageencoder}$-dimensional output embeddings are mapped to the LLM’s input dimension using a projection network ($MLP\!: d_{\imageencoder} \to d_{\llm}$). Further, to compress visual tokens, spatial $D\times D$ down-sampling is applied to each tile's patch embeddings (typically $2\times2$). A tile is ultimately represented by $N\!=\!\tfrac{HW}{D^2P^2}$ visual tokens (typically $N{=}256$). Although this spatial downsampling introduces an information bottleneck, it remains essential for maintaining computational efficiency. Visual tokens are then concatenated to form $\mathbf{V} = [V_1, \ldots, V_N]$, which is then fed to the LLM. For multiple tiles i.e. $T>1$, visual tokens from all tiles are concatenated to form $\mathbf{V}$.

In parallel, the LLM’s text decoder embeds the query $q$ into $M$ tokens of dimension $d_{\llm}$, denoted as $\mathbf{T} = [T_1, \ldots, T_M]$.

The core layers of the LLM then receive a concatenation of both sets of embeddings as input and the VLM generates the answer conditioned on the input embeddings:
\begin{align}
\hat{a}_i \sim P_{\mathrm{VLM}}\!\left(a_i \mid \hat{a}_{<i};  \mathbf{V}, \mathbf{T}\right).
\label{eq:gen}
\end{align}

\section{Methodology}\label{sec:methodology}
Addressing the fundamental limitation of encoding an image irrespective of the task at hand, we present a \textit{Text-guided Image Encoder} (\tie).
\begin{figure*}[ht]
    \centering
    \includegraphics[width=\linewidth]{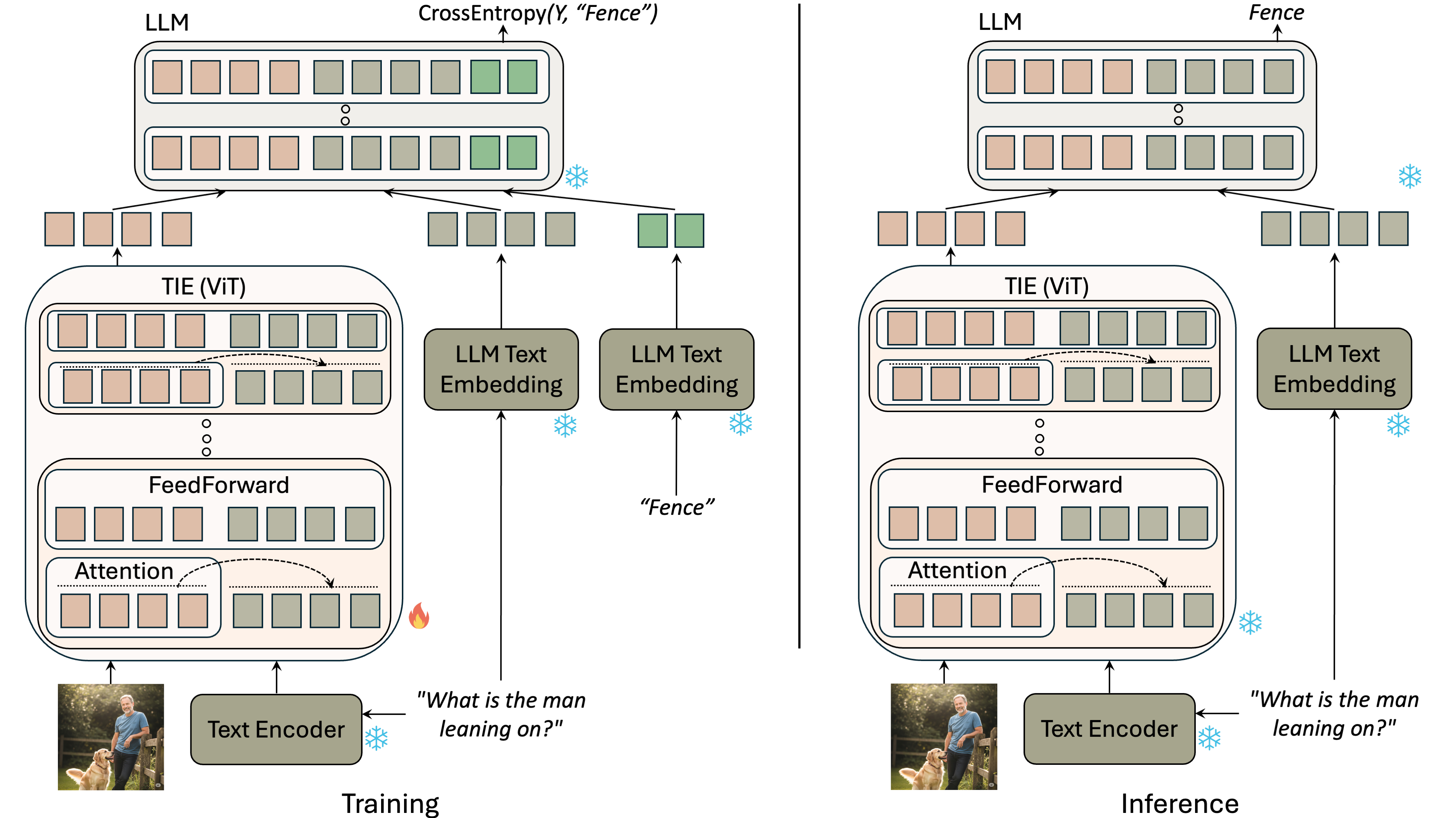}
    \caption{\tie encodes an image conditioned on the corresponding query, yielding semantically enriched, query-specific image representations. Conditioning is performed across all layers of \tie. \ours training mechanism is depicted on the left, and inference on the right.
    }
    \label{fig:main}
\end{figure*}
At its core, the \tie model leverages the instruction/query of interest, denoted by $q$, to enhance the quality of image encoding for a given image tile $I$. 
In a standard Vision Transformer (ViT), the input $N_\imageencoder$ patches are processed with a patch embedder (typically a 2D-convolutional layer), then
rotary position embeddings (RoPE) are incorporated into the patch-level feature maps to maintain spatial coherence within the image representation. 
We will denote by $[I_{1}, I_{2}, \ldots, I_{N_\imageencoder}]$ this sequence of position-augmented patch embeddings.

In the TIE framework, we augment this process by allowing the ViT to \emph{attend} to the query $q$ during encoding, thereby enabling the resulting image representations to be conditioned on the query of interest. 
Inspired by \citet{zha2025language}, TIE employs a pretrained text encoder (e.g., T5), to encode the query $q$ into stable and contextually rich textual embeddings. 
These text representations are then integrated into the image encoder’s attention mechanism such that visual patches can attend to the query tokens. Let $\mathbf Q$ be the sequence of embeddings from the text encoder: 
\begin{align}
\mathbf{Q} = [Q_1,...Q_L] &= \textencoder(q).
\end{align}
Since these embeddings originate from a transformer model that already incorporates positional encodings, we omit the application of RoPE to the query embeddings.

The query embeddings are concatenated with the image embeddings and are supplied to the image encoder. 
\begin{align}
U = \tie(I, q) = \imageencoder([I_{1}, I_{2},....I_{N_\imageencoder}, Q_1,...Q_L]). \label{eq:7}
\end{align}
The attention pattern is set such that the image tokens can attend to text tokens but not the other way around. While the text tokens are not modified in the attention module,
they still pass through the feedforward network in layers of the ViT transformer~\citep{dosovitskiy2020image}. Residual connections across the attention and feedforward networks are kept unchanged. Since the \textencoder representations are already contextual, we did not update them further in the attention module, as preliminary experiments showed this yields better results.

We extract the first $N_\imageencoder$ tokens from $U$ corresponding to the $N_\imageencoder$ patches, and use them as the final output of the augmented image encoder $\tie$. As with the standard \imageencoder, these embeddings are further projected using $MLP$ and spatially down-sampled to $N$ tokens $\mathbf V=[V_1, \ldots,V_N]$.

For the multi-tile setting ($T>1$), \tie follows the standard image-encoder workflow: each tile is encoded independently, and the resulting $N$ tokens per tile are all concatenated to form $\mathbf{V}$.
The rest of the LLM processing follows the steps described above with the augmented set of visual tokens $\mathbf V$ fed to the VLM of \Cref{eq:gen}. 
\Cref{fig:main} illustrates the overall methodology, detailing the processes followed during both training and inference stages.

\subsection{Training the \tie\ Model}\label{ss:training}
We adopt the same \textit{language alignment} procedure as PLM to train the \tie\ model. Starting with a pretrained and already aligned image encoder, we integrate an additional text encoder into the image encoder architecture and perform language alignment training while keeping only the image encoder parameters trainable. During language alignment, the model is trained using the standard \textit{language modeling} objective, where the cross-entropy loss is minimized between the predicted token distribution from the VLM and the ground-truth answer sequence $a^*$. Formally,
\begin{equation}
\mathcal{L}_{\text{align}} = - \sum_{i=1}^{N} \log P_{\mathrm{VLM}}\!\left(a_i^* \mid a_{<i}^*; \mathbf{V}, \mathbf{T}\right),
\end{equation}
In our experiments, we initialize the image encoder ($IE$), the LLM and the attached downsampling projector with a pretrained VLM model, e.g, \plm~\citep{cho2025perceptionlm}. Since this work focuses on image encoders, we fine-tune only the image encoder and the projection $MLP$, keeping the LLM and text encoder within \tie frozen as shown in \Cref{fig:main}.

\section{Experiments}
\begin{table*}[ht!]
\centering
\resizebox{\textwidth}{!}{%
\begin{tabular}{l|c|c|c|rrrrrrrr|r}
\toprule
\textbf{Model} & \textbf{Enc.} & \textbf{Dec.} & $T$ & \textbf{DocVQA} & \textbf{ChartQA} & \textbf{TextVQA} & \textbf{InfoVQA} & \textbf{OCRB} & \textbf{OKVQA} & \textbf{VIZWIZ} & \textbf{MME}  & \textbf{Avg.} \\
\midrule
\textsc{Qwen2VL-2B}$^{\dagger}$  &$0.67$B&$1.5$B& Dynamic\footnotemark[1]& 90.1 & 75.3 & 80.3 & 65.5 & 80.9 & 59.7 & 67.4 & 69.29 & 73.56 \\
\textsc{InternVL2.5-1B}$^{\dagger}$ &$0.3$B&$0.5$B&up to 36& 84.8 & 75.9 & 72.0 & 56.0 & 78.5 & 51.5 & 47.4 & 69.64 & 66.97 \\
\plm-\textsc{1B}$^{\dagger}$ &$0.3$B&$1$B&up to 36& 90.7 & 78.6 & 82.1 & 63.0 & 80.7 & 61.0 & 59.7 & 57.25 & 71.63 \\\hline
\textsc{Qwen2.5VL-3B}$^{\dagger}$ &$0.67$B&$3$B & Dynamic\footnotemark[1]& 93.9 & 83.1 & 79.3 & 77.1 & 79.7 & 63.2 & 71.9 & 79.32 & 78.44 \\
\textsc{InternVL2.5-4B}$^{\dagger}$ &$0.3$B&$0.5$B&up to 36& 91.6 & 84.0 & 79.3 & 72.1 & 52.3 & 64.0 & 61.8 & 83.46 & 73.57 \\
\plm-\textsc{3B}$^{\dagger}$ &$0.3$B&$3$B&up to 36& 93.8 & 84.3 & 84.3 & 74.6 & 83.0 & 66.8 & 64.0 & 67.11 & 77.24 \\
\midrule\midrule
\plm-\textsc{1B}$^\ddagger$ & \multirow{3}{*}{\shortstack{$0.3$B\\(448/14)}} & \multirow{3}{*}{$1$B} & \multirow{3}{*}{$1$} & 51.05 & 58.56 & 65.34 & 27.25 &  59.20 & 55.76 & \textbf{59.18} & 54.18 & 53.82 \\
\bslneC-\textsc{1B} & & & & 68.61 & 72.28 & 72.57 & 33.92 & 64.70 & \textbf{56.02} & 57.14 & 58.7 & 60.49 \\
\ours-\textsc{1B} & & & &\textbf{73.92} & \textbf{73.32} & \textbf{72.57} & \textbf{37.32} & \textbf{66.40} & 55.90 & 57.61 & \textbf{59.11} & \textbf{62.02} (\textbf{+1.5}) \\
\midrule
\plm-\textsc{3B}$^\ddagger$ & \multirow{3}{*}{\shortstack{$0.3$B\\(448/14)}} & \multirow{3}{*}{$3$B} & \multirow{3}{*}{$1$} &52.83 & 60.48 & 66.42 & 31.84 & 61.00 & 61.51 & 59.20 & 64.01 & 57.16 \\
\bslneC-\textsc{3B} & & & & 72.61 & 76.68 & 73.60 & 38.02 & \textbf{68.60} & 59.37 & 59.19 & 65.93 & 64.25 \\
\ours-\textsc{3B} & & & & \textbf{75.42} & \textbf{76.80} & \textbf{74.28} & \textbf{40.75} & 66.80 & \textbf{61.01} & \textbf{63.28} & \textbf{65.99} & \textbf{65.54} (\textbf{+1.3}) \\
\bottomrule
\end{tabular}}%
\caption{Evaluation results on nine benchmarks. 
For the evaluated models (last six rows), the number of tiles, $T=1$. 
$Enc.$ denotes image encoder’s size and resolution and $Dec.$ denotes the LLM size.
\ours consistently outperforms both \bslneC and \plm across $1$B, $3$B sizes. $^{\dagger}$-reported in the respective works, \plm~\citep{cho2025perceptionlm}. $^{\ddagger}$-\plm trained with 36 tiles (1 tile training during initial stages) evaluated at 1 tile.
}\label{tab:model_sizes}
\end{table*}

\subsection{Model and Dataset Details}
\paragraph{Training Details.} 
We train our image encoder on top of pretrained \pef (\pes) model~\citep{bolya2025perception}. \citet{cho2025perceptionlm} used $\pes{-}\rm L$ (0.3B params) as the image encoder and released \plm models of varying scales.
$\pes{-}\rm L$ has a patch size of $P=14$ and a tile size of $H\times W=448\times 448$. The \plm projector applies a spatial downsampling of $D=2$, resulting in 256 tokens per tile. 
In our experiments, the \tfl encoder \citep{raffel2020exploring} (0.3B params) was used as the auxiliary text encoder in \tie. The \tie image encoder with the \plm language model as a whole is referred to as \ours. In each case, the \tie model is trained with a peak learning rate of $1e^{-4}$ with weight decay of $0.01$ for $25K$ steps with an effective batch size of $2048$.
In most cases (except for \Cref{tab:multi_tiles}), both training and inference are conducted with a single tile, gi.e., $T{=}1$. Additional details in \Cref{sec:appendix}.

\paragraph{Data Details.}
We utilize the stage-$2$ training dataset curated by \citet{cho2025perceptionlm}, using only the image samples and excluding any video and text-only data. For simplicity, we retain only the first turn from multi-turn samples and conduct training on single-image datapoints. Nonetheless, the proposed approach can be easily modified to accommodate multi-turn and interleaved image inputs. The final dataset comprises 37M datapoints each with one image.

\paragraph{Models Compared.} 
\plm models of varying sizes ($1$B, $3$B) serve as the primary baselines for comparison. \plm\ models were trained with varying tile per image (1 through 16) across different training stages, followed by final tuning with 36 tiles. Trained \plm models are evaluated at different tile settings during inference. To fairly assess the compute involved in training \ours\ (\Cref{ss:training}), we construct a \bslneC variant that continues pretraining the \plm model on the same data (identical batches) and for the same number of optimization steps as the \ours model. Notably, the image encoder in \ours has the same number of trainable parameters as the corresponding image encoder in \bslneC. Parameters in \tfl are kept frozen as discussed earlier. Note that \ours uses both an instruction prompt - ``\textit{Answer the following question about the given image:}’’ followed by the corresponding query as the input to \tie.

\begin{wrapfigure}{r}{0.52\textwidth}
    \centering
    \includegraphics[width=\linewidth]{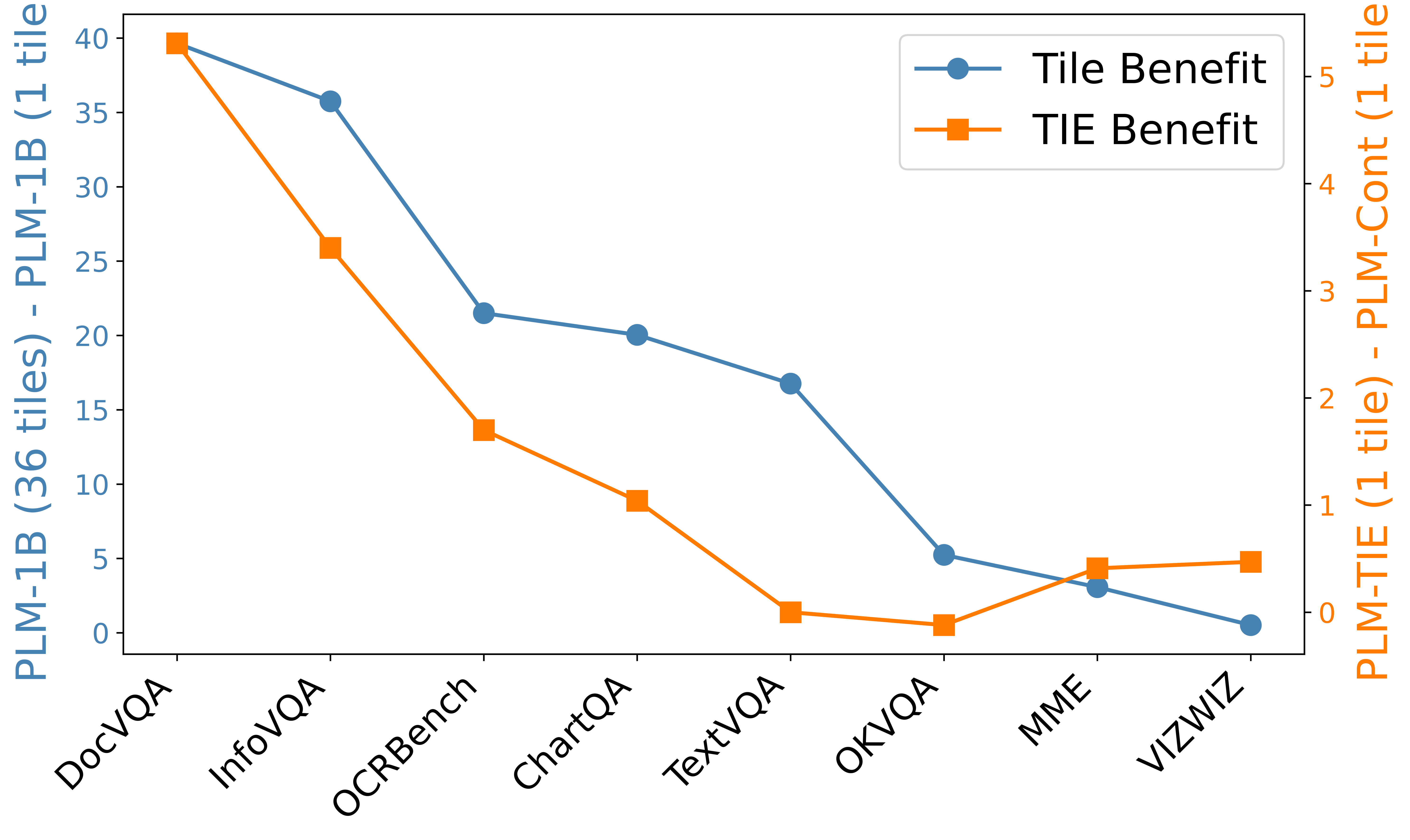}
    \caption{\plm-\textsc{1B} performance difference between 36 and 1 tiles, compared with the \tie{}–baseline gap. On datasets that benefit from more tiles, \tie correspondingly achieves larger gains. 
    }\label{fig:diff}
\end{wrapfigure}

\paragraph{Evaluation Benchmarks.}
We compare \ours with \bslneC and \plm over 14 diverse image-to-text tasks spanning OCR, chart, document, and visual QA, as evaluated in~\citet{bolya2025perception}. The tasks include – DocVQA~\citep{mathew2021docvqa}, 
ChartQA \citep{zheng2025advancing}, 
TextVQA \citep{singh2019towards},
InfoVQA \citep{mathew2022infographicvqa}, 
AI2D \citep{kembhavi2016diagram}, 
OCRBench \citep{liu2024ocrbench}, 
MMMU \citep{yue2024mmmu}, 
OK-VQA \citep{schwenk2022okvqa}, 
VizWiz \citep{bigham2010vizwiz}, 
MME \citep{chaoyou2023mme}, 
POPE \citep{li2023evaluating}, 
Flickr \citep{young2014image}, 
COCO \citep{lin2014microsoft}, 
NoCap \citep{agrawal2019nocaps}. 
We adopt evaluation protocols consistent with \plm and those followed by LMMS-Eval \citep{zhang2024lmmsevalrealitycheckevaluation, lmms_eval2024}.
OCRBench and MME scores are further normalized by 1000 (\#datapoints) and 2800 (maximum score), respectively, to bring them to a similar scale (out of 100) for averaging with other datasets. 

\subsection{Main Results}
At the 1B and 3B scales, \ours consistently outperforms both \bslneC and \plm, achieving average gains of +1.5 and +1.3 points over \bslneC, resp. (\Cref{tab:model_sizes}). This demonstrates that the \tie image encoder is able to create enhanced image representations utilizing the query at hand. 

\Cref{fig:diff} shows the \plm-\textsc{1B} performance gap between $1$ and $36$ tiles, relative to the \ours and \bslneC gap at $1$ tile. On datasets where \plm benefits from using a larger number of tiles -  DocVQA, InfoVQA, ChartQA, etc. \ours outperforms \bslneC by up to $5$ points. 
Given the computational cost of training and inference grows sharply with tile count, \tie thus provides an orthogonal route to performance gains without relying on additional tiles. 

On other VQA datasets—AI2D, MMMU, and POPE, \ours performs comparably to \bslneC, showing only modest gains. We attribute this to findings by \citet{tong2024cambrian}, who demonstrated that textual inputs (queries) have a greater influence than visual inputs on these datasets. Results for these datasets are presented in \Cref{tab:model_sizes_full} of \Cref{ap:additional_datasets}. 
Captioning dataset results - COCO, NOCAPS, FLICKR30K are also shown in \Cref{tab:coco_nocaps_flickr} of \Cref{appsub:captioning_data}. Since these datasets lack example-specific queries and the training split includes relatively less captioning data, \ours gains limited benefit from text-conditioning and thus performs comparably to \bslneC on captioning benchmarks.\footnotetext[1]{Qwen uses naive dynamic resolution i.e. dynamic number of tokens}

\begin{table*}[!h]
\centering
\scriptsize
\resizebox{\textwidth}{!}{%
\begin{tabular}{l|c|c|rrrrrrrr|r}
\toprule
\textbf{MultiTile} & \textbf{Dec.} & $T$ & \textbf{DocVQA} & \textbf{ChartQA} & \textbf{TextVQA} & \textbf{InfoVQA} & \textbf{OCRB} & \textbf{OKVQA} & \textbf{VIZWIZ} & \textbf{MME} & \textbf{Avg.} \\
\midrule
\plm-\textsc{1B} & \multirow{3}{*}{$1$B} & \multirow{3}{*}{up to $2$} & 54.44 & 61.08 & 66.05 & 36.81 & 62.20 & 55.29 & \textbf{59.14} & 56.16 &	56.40 \\
\bslneC-\textsc{1B} & & & 66.67 & 72.04 & 71.44 & 38.41 & 65.90 & 56.56 & 57.08 & 57.47 & 60.70 \\
\ours-\textsc{1B} & & & \textbf{71.78} & \textbf{73.36} & \textbf{71.56} & \textbf{43.96} & \textbf{66.90} & \textbf{56.81} & 57.43 & \textbf{58.16} & \textbf{62.50} (\textbf{+1.8}) \\
\midrule
\plm-\textsc{1B} & \multirow{3}{*}{$1$B} & \multirow{3}{*}{up to $4$} & 80.92 & 72.44 & 77.59 & 50.16 & 73.50 & 55.83 & \textbf{59.78} & 58.59&	66.10 \\
\bslneC-\textsc{1B} & & & 83.70 & 76.92 & \textbf{80.78} & 48.70 &\textbf{74.40} & 57.43 & 57.80 & 58.30 & 67.25 \\
\ours-\textsc{1B} & & & \textbf{86.72} & \textbf{77.68} & 80.50 & \textbf{57.09} & 73.70 & \textbf{57.63} & 56.69 & \textbf{60.15} & \textbf{68.77} (\textbf{+1.5}) \\
\midrule
\plm-\textsc{1B} & \multirow{3}{*}{$1$B} & \multirow{3}{*}{up to $8$} & 81.30 & 74.80 & 79.79 & 54.66 & \textbf{74.70} & 55.88 & \textbf{59.66} & 57.61&	67.30 \\
\bslneC-\textsc{1B} & & & 84.11 & 78.44 & \textbf{81.49} & 52.09 & 73.40 & \textbf{57.69} & 57.60 & 58.23 & 67.88 \\
\ours-\textsc{1B} & & & \textbf{86.88} & \textbf{78.56} & 80.76 & \textbf{58.32} & 73.90 & 57.13 & 56.64 & \textbf{58.30} & \textbf{68.81} (\textbf{+0.9}) \\
\bottomrule
\end{tabular}}
\caption{Evaluation results across benchmarks for 2-, 4-, and 8-tile at 1B scale. \ours outperforms \bslneC and \plm across the settings. Also, \tie($T=4$) beats \bslneC($T=8$). \ours-\textsc{1B} ($T=8$) gets close to the \plm-\textsc{1B} ($T=36$) version from \Cref{tab:model_sizes}.}\label{tab:multi_tiles}
\end{table*}

\subsection{Multiple tiles}
While \tie improves performance without extra tiles, increasing tile count remains valuable as it alleviates the image encoder bottleneck (\Cref{sec:background}) and can yield additional performance gains. We evaluate the proposed \tie encoder under varying tile counts in \Cref{tab:multi_tiles}. Specifically, we experiment with three settings where the maximum number of tiles is constrained to $2$, $4$, and $8$ with $1$B sized models. \ours and \bslneC were both trained and evaluated with this setting. Pretrained \plm evaluated at each given $T$ is also included for comparison. Within \tie, each tile of the image is independently encoded by the \tie encoder, conditioned on the query of interest (\Cref{sec:methodology}). 

To derive image tiles, we follow the approach of MetaCLIP~\citep{chuang2025meta} and \plm, selecting the largest possible image that fits within the specified tile limit while preserving the original aspect ratio. Across all three evaluated settings, \ours consistently outperforms the baseline, demonstrating that the proposed \tie encoder effectively handles multi-tile inputs and achieves superior performance.

Note that the performance gains at $T=2$ are larger than those at $T=4.8$, which is expected since the bottleneck effect diminishes as the number of tiles increases. Notably, \ours with 4 tiles exceeds \bslneC at 8 tiles, underscoring the efficiency of the architecture. 

\begin{table}[t]
\centering
\small
\resizebox{0.6\textwidth}{!}{%
\begin{tabular}{l|r|rr|r}
\toprule
\textbf{Model} & \textbf{Memory (GB)} & \multicolumn{2}{c|}{\textbf{Time}} & \textbf{Avg. Score} \\
\cline{3-4}
 &  & \textbf{Training (h)} & \textbf{Inference (s)} &  \\
\midrule
\bslneC-\textsc{1B} & 3.48 & \textbf{14.5} & \textbf{0.15} & 60.49 \\
\ours-\textsc{1B} & 4.12 & 17.5 & 0.16 & \textbf{62.02} \\\hline
\bslneC-\textsc{1B} ($T{=}8$) & 4.48 & \textbf{44} & 0.24 & 67.88 \\
\ours-\textsc{1B} ($T{=}4$) & 4.62 & \textbf{44} & \textbf{0.21} & \textbf{68.77} \\
\bottomrule
\end{tabular}}
\caption{Comparison of memory usage, runtime, and performance. \ours ($T{=}4$) beats \bslneC ($T{=}8$) while being much faster in inference and matching training time.}\label{tab:runtime_memory}
\end{table}

\subsection{\tie's Runtime \& Memory footprint}

Having established \tie's superior accuracy, we now analyze its efficiency across two key dimensions. First, we evaluate its runtime and memory footprint. Second, we demonstrate its strong performance in token-constrained settings (\Cref{ss:fewer_tokens}).

\Cref{tab:runtime_memory} reports the training time (for 25k steps), peak inference-time memory usage and per-example inference runtime across different models. Training runtime is reported on the entire training data. 
More details on compute used are mentioned in \Cref{apsec:resources}. For inference comparisons, since LLM runtime is typically dominated by long generations, we focus our measurements on tasks with shorter output lengths to more accurately capture the runtime overhead introduced by the image encoder. Notably, for tasks involving longer generations, the runtime difference narrows further, making \ours even more comparable to \bslneC. Here, we focus specifically on the OCRBench task, where the average answer length is $<5$ tokens—relatively short compared to other benchmarks.

As expected, \ours exhibits a slightly higher memory footprint and runtime due to the additional text encoder in \tie. Nevertheless, as seen from the first two rows, this overhead is accompanied by a substantial improvement in performance. Furthermore, comparing \ours ($T{=}4$) with \bslneC ($T{=}8$) shows that \ours delivers higher accuracy with fewer tiles, while also being faster at inference and matching training time. Hence, the proposed \tie architecture achieves both efficiency and performance gains. Although memory usage remains slightly higher, later in Ablation 1 we introduce versions of \ours\ that match the memory footprint of \bslneC.

\subsection{Fewer Tokens per Image}\label{ss:fewer_tokens}

\begin{figure}[t]
    \centering
    \begin{subfigure}[b]{0.48\linewidth}
    \includegraphics[width=\linewidth]{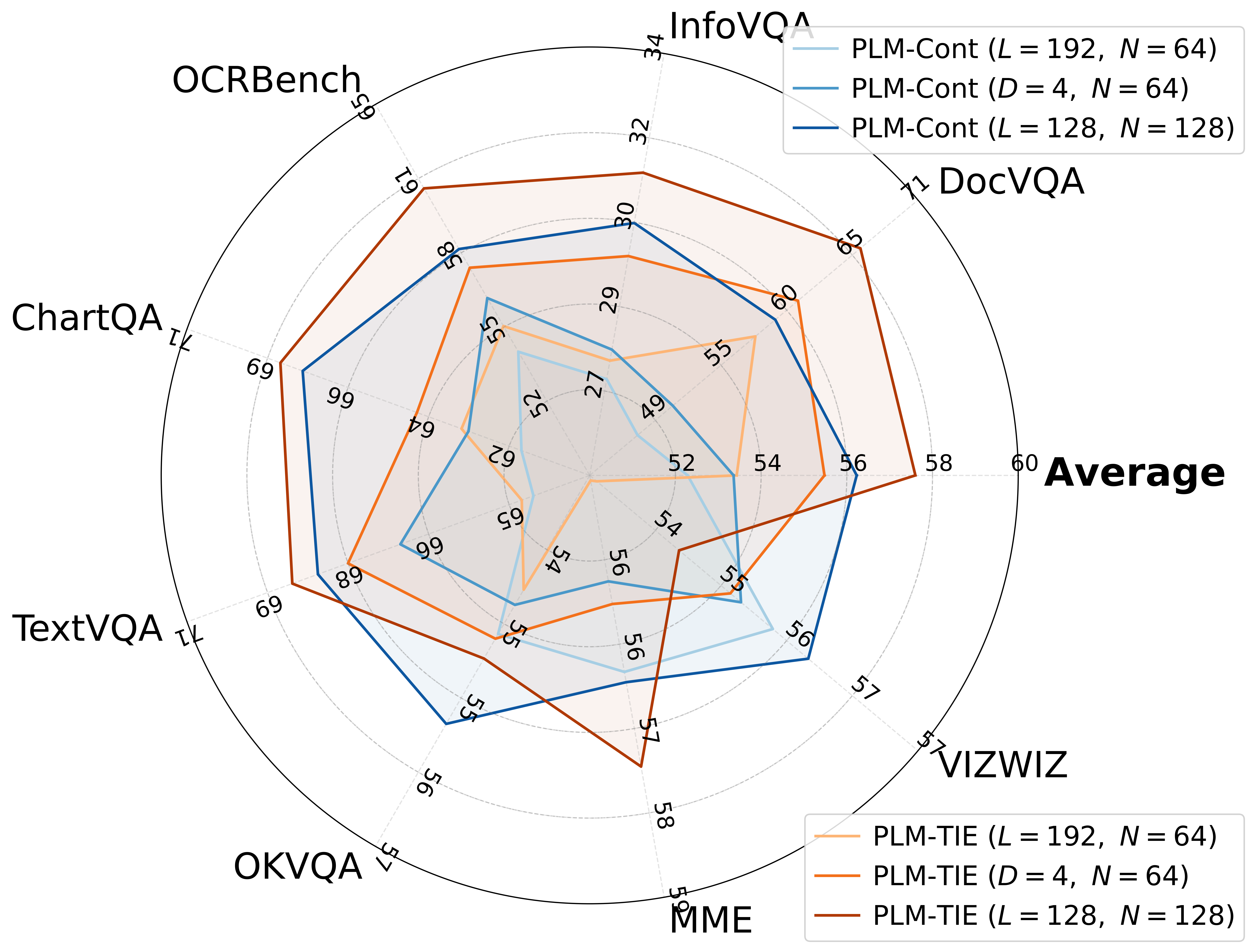}
    \caption{Models compared with fewer tokens per tile. \ours does better than the corresponding \bslneC across fewer token setups (larger $D$; larger $L$).
    }
    \label{fig:poolvsL}
    \end{subfigure}\hfill
    \begin{subfigure}[b]{0.48\linewidth}
    \includegraphics[width=\linewidth]{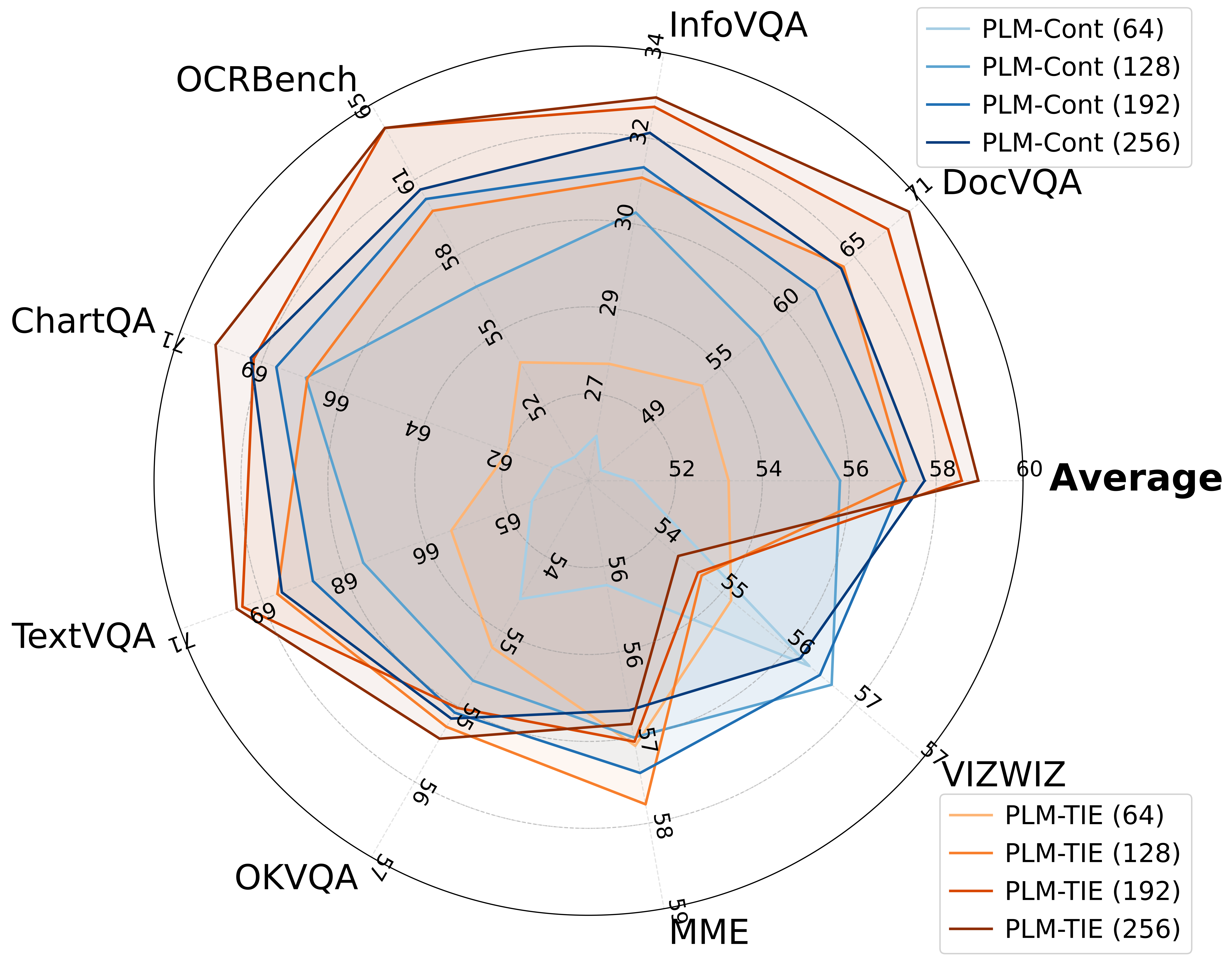}
    \caption{Matryoshka-style 1D models trained with $L \in {64,128,192,256}$ sampled per batch, and evaluated at fixed a $L$. \ours $(192)$ beats \bslneC $(256)$. 
    }
    \label{fig:ftpt}
    \end{subfigure}
    \caption{Ablation results. Fewer tokens per image, either with dedicated models (left) or versatile models than support a variable number of tokens (right).}
\end{figure}

Using fewer tokens per image significantly boosts efficiency, which is critical for video, large images, or processing many images. 
We modify the model architecture to represent a tile using a smaller number of tokens.
The simplest way to achieve this is by using a larger downsampling rate $D=4$ resulting in $N\!=\!\tfrac{HW}{D^2P^2}\!=\!64$ tokens.
Alternatively, recent works in image generation~\citep{yu2024image, yan2024elastictok, zha2025language} have proposed training models in a 1D token format, where the first $N\!=\!\tfrac{HW}{D^2P^2}\!-\!L$ output tokens from the image encoder are optimized for downstream tasks. While smaller changes in $D$ change $N$ significantly because of the squared dependence in the first case, the dependence is less constrained with $L$ in 1D.
\Cref{fig:poolvsL} shows the models trained and evaluated with fewer tokens per tile using both larger downsampling and 1D. $D=2$ was used with 1D models. $T\!=\!1$ in both cases. As shown, \ours outperforms \bslneC in both setups.  

\begin{table*}[ht!]
\centering
\resizebox{\textwidth}{!}{%
\begin{tabular}{ll|rrrrrrrr|r}
\toprule
\textbf{Abl.} & \textbf{Model} & \textbf{DocVQA} & \textbf{ChartQA} & \textbf{TextVQA} & \textbf{InfoVQA} & \textbf{OCRB} & \textbf{OKVQA} & \textbf{VIZWIZ} & \textbf{MME} & \textbf{Avg.} \\\midrule
\multirow{3}{*}{1} &
\bslneC-\textsc{1B} & 68.61 & 72.28 & 72.57 & 33.92 & 64.70 & 56.02 & 57.14 & 58.7& 60.49 \\
& \ours-\textsc{1B} (\texttt{with VLM's TE}) & 72.43 & 72.80 & 72.71 & 36.57 & 65.30 & 56.62 & 55.30 & 59.65&	\underline{61.42} \\
& \ours-\textsc{1B} (with \tfl) & 73.92 & 73.32 & 72.57 & 37.32 & 66.40 & 55.90 & 57.61 & 59.11&	\textbf{62.02} \\
\midrule
\multirow{3}{*}{2} & \ours-\textsc{1B} & 73.92 & 73.32 & 72.57 & 37.32 & 66.40 & 55.90 & 57.61 & 59.11&	\textbf{62.02} \\
& \ours-\textsc{1B} (\texttt{Generic Query @Inference}) & 71.33 & 73.32 & 71.99 & 35.81 & 65.80 & 55.97 & 58.69 & 59.39&	\underline{61.54} \\
& \ours-\textsc{1B} (\texttt{Instruction only @Inference}) & 68.92 & 72.88 & 70.41 & 33.80 & 64.70 & 55.94 & 58.25 & 58.94&	60.48
  \\\midrule
\multirow{4}{*}{3}  & \ours-\textsc{1B} & 73.92 & 73.32 & 72.57 & 37.32 & 66.40 & 55.90 & 57.61 & 59.11&	\underline{62.02} \\
 & \ours-\textsc{1B} (\texttt{Question only})& 73.40 & 72.60 & 73.03 & 36.91 & 66.30 & 56.85 & 57.52 & 59.84&	\textbf{62.06} \\
& \ours-\textsc{1B} (\texttt{Instruction only}) & 68.90 & 72.48 & 72.46 & 33.56 & 64.80 & 56.72 & 57.53 & 57.13&	60.45 \\
& \ours-\textsc{1B} (\texttt{Dummy Question}) & 67.32 & 71.84 & 72.18 & 34.56 & 66.30 & 56.40 & 55.75 & 57.93&	60.29 \\\midrule 
\multirow{3}{*}{4} & \ours-\textsc{1B} (\texttt{T5 Same}) & 73.69 & 72.68 & 73.05 & 37.90 &  65.80 & 56.32 & 57.44 & 58.07&	61.87 \\
& \ours-\textsc{1B} & 73.92 & 73.32 & 72.57 & 37.32 & 66.40 & 55.90 & 57.61 & 59.11&	\underline{62.02} \\
& \ours-\textsc{1B} (\texttt{Cross Encoder}) & 74.25 & 73.12 & 72.94 & 37.31 & 69.10 & 56.58 & 57.02 & 58.57&	\textbf{62.36} \\
\bottomrule
\end{tabular}}
\caption{Ablation studies. (1) Using VLM's non-contextual embeddings already results in notable boost over the baseline. (2)\ours when used with a generic question at test time performs better than \bslneC. (3) Merely increasing compute inside the image encoder is not sufficient. Question conditioning is important. (4) Adding more parameters to the \tie image encoder might be beneficial.}\label{tab:ablations}
\end{table*}

The key advantage of 1D models lies in their ability to operate in a Matryoshka-style manner \citep{kusupati2022matryoshka}, handling a variable number of tokens $N$ per image. These models are trained by randomly varying $L$ value during training enabling it to adapt to any $L$ at inference depending on the setting and images. To evaluate \tie for such setups, we train a single model randomly varying $L$ in $[192,128,64,0]$. We evaluate this model with all the four $L$ values for both \bslneC and \ours. \Cref{fig:ftpt} shows \ours outperforms \bslneC across the $L$ values. Thus, the experiments show that \ours outperforms \bslneC with fewer tokens per image.

\subsection{Ablations/Analysis}
In this subsection, we analyze the key components that contribute to \tie’s performance and examine its adaptability across varying configurations and use cases. All ablation results and analyses are summarized in \Cref{tab:ablations}. 

\subsubsection{Towards Deeper, Context-Aware Image Encoders}
We examine here the effectiveness of a VLM’s internal text encoder in generating text-conditioned image representations. We replace the \tfl component in \tie with the default text encoder from the VLM. The results from ablation~1 reveal a striking finding: 
Even when using only the VLM’s non-contextual input-layer token embeddings, performance already surpasses the \bslneC baseline by a notable margin of $+0.9$.
This suggests that even the initial token embeddings of the language model encode meaningful semantic structure that benefits image embeddings. Future work could explore leveraging deeper, contextually enriched representations from the VLM to further enhance the image encoder’s alignment and semantic grounding.

\begin{figure*}[ht!]
    \centering
    \begin{subfigure}[b]{0.48\textwidth}
        \centering
        \begin{subfigure}[b]{0.49\textwidth}
            \centering
            \includegraphics[width=\linewidth]{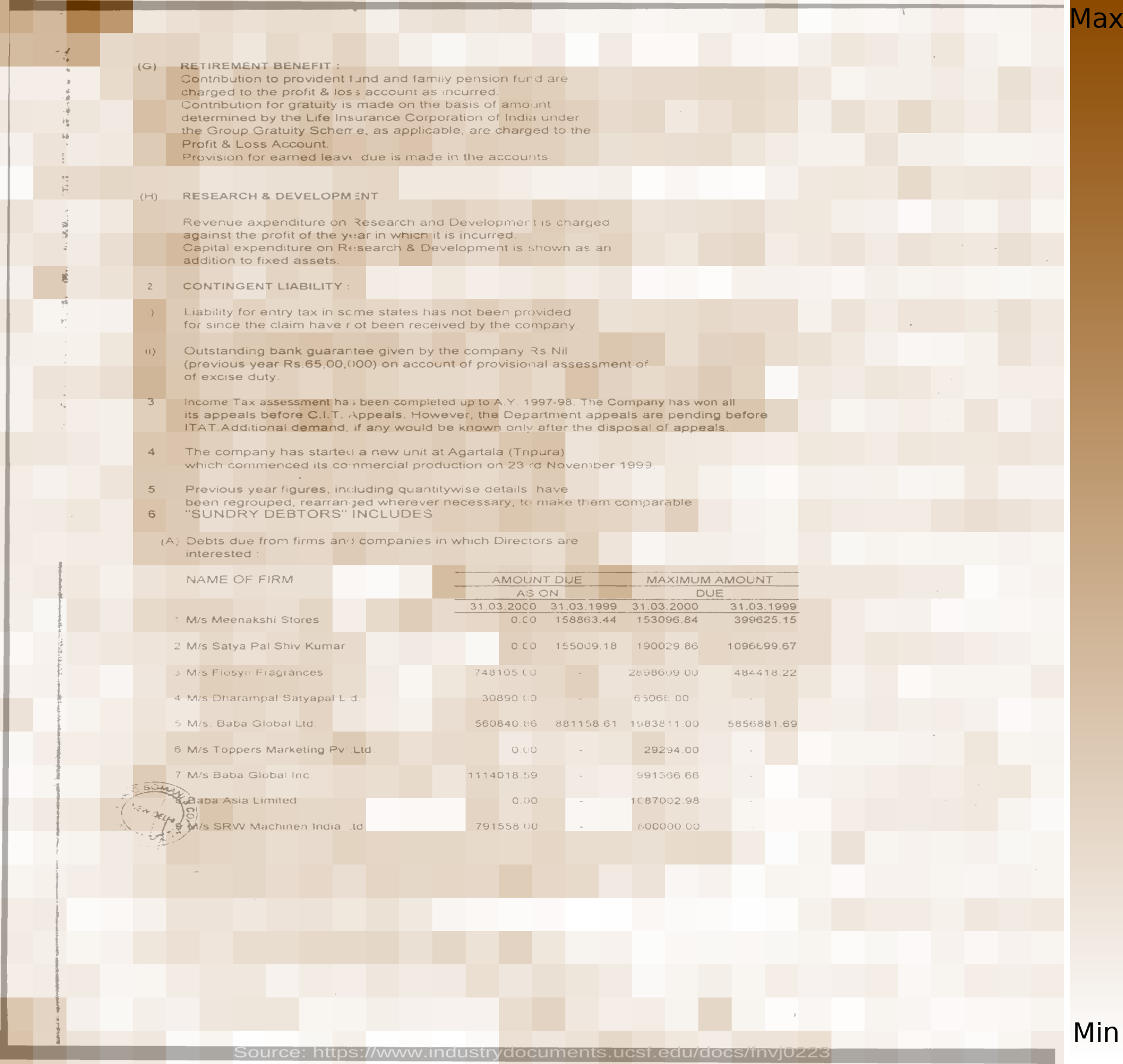}
        \end{subfigure}
        \hfill
        \begin{subfigure}[b]{0.49\textwidth}
            \centering
            \includegraphics[width=\linewidth]{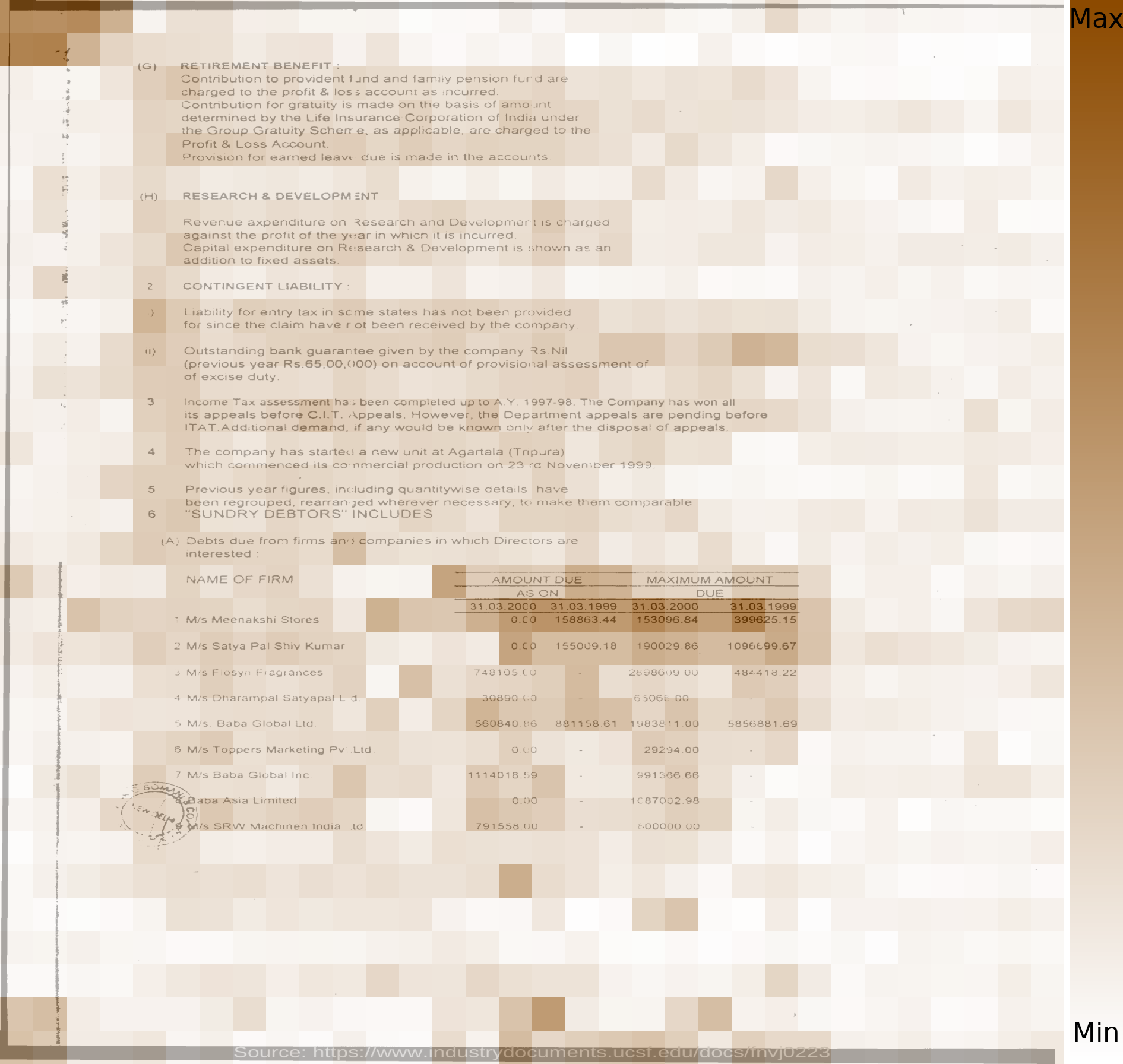}
        \end{subfigure}
        \caption{Question is \textit{``How much is the amount due of "M/s Meenakshi Stores" as on 31.03.1999 based on section 6 "SUNDRY DEBTORS"''.}}
        \label{fig:fig5.a}
    \end{subfigure}
    \hfill
    \begin{subfigure}[b]{0.48\textwidth}
        \centering
        \begin{subfigure}[b]{0.49\textwidth}
            \centering
            \includegraphics[width=\linewidth]{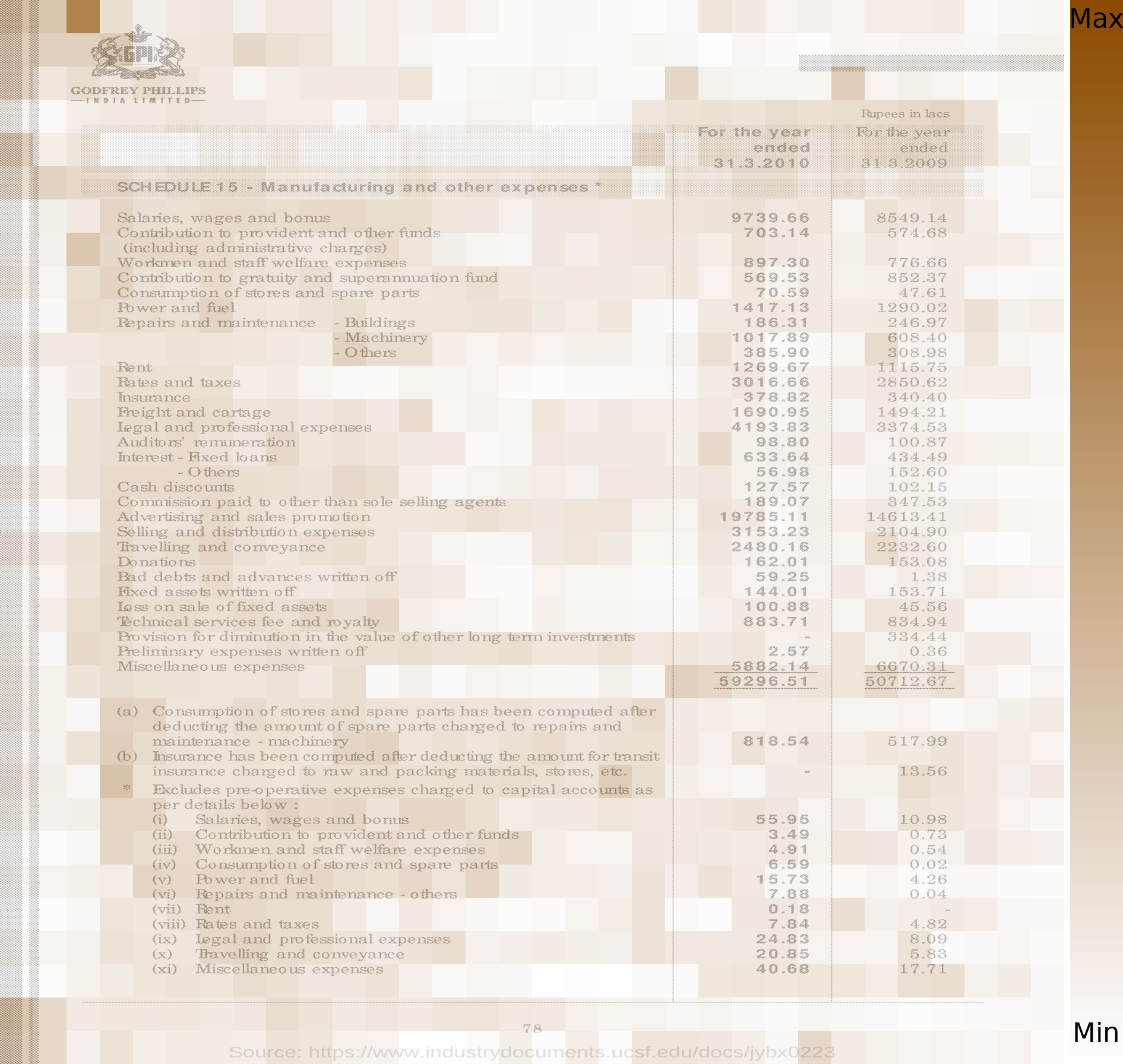}
        \end{subfigure}
        \hfill
        \begin{subfigure}[b]{0.49\textwidth}
            \centering
            \includegraphics[width=\linewidth]{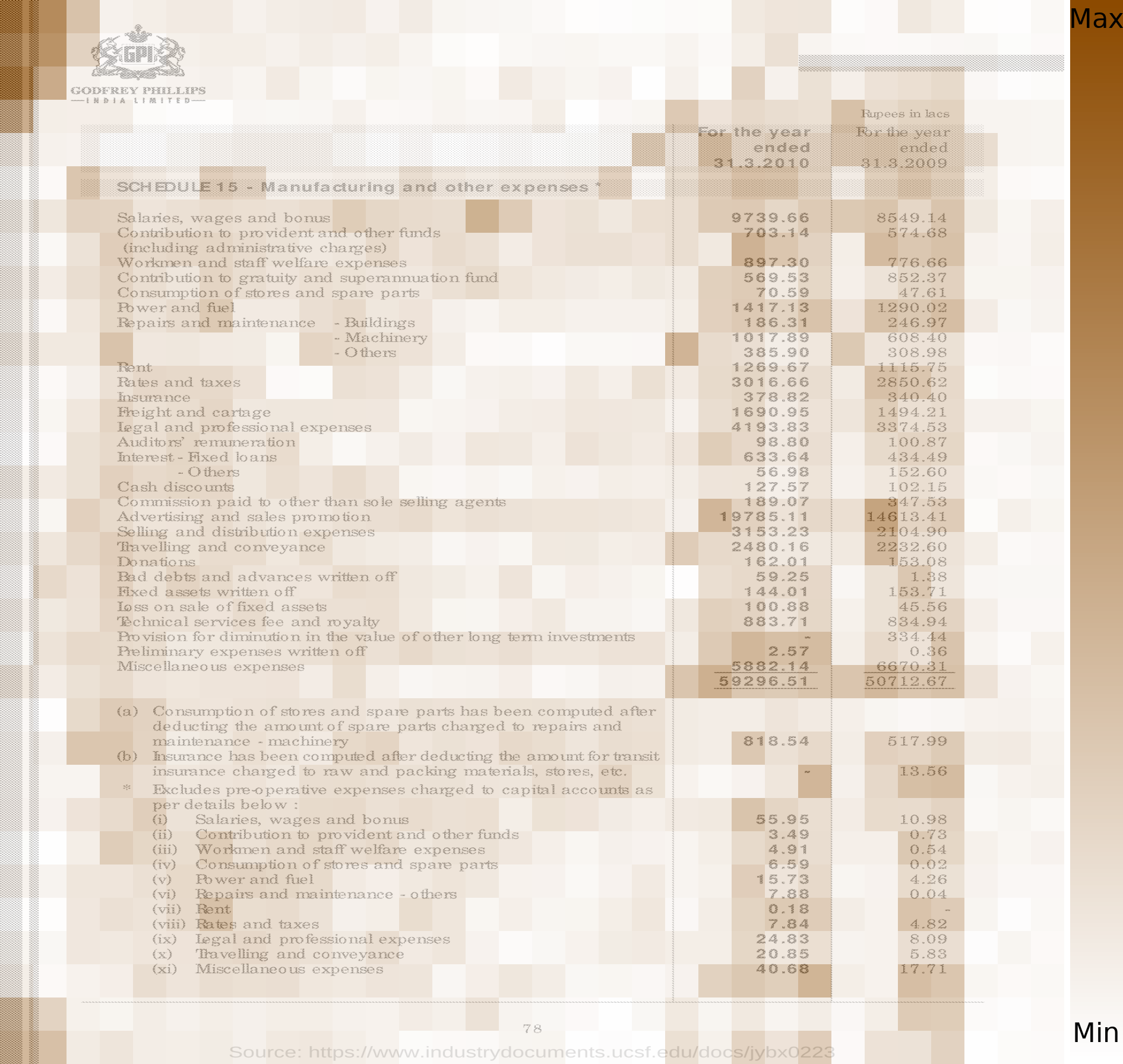}
        \end{subfigure}
        \caption{Query is \textit{``How much is the "contribution to provident and other funds (including administrative charges)" expenses for the year ended 31.3.2009''.}}
        \label{fig:fig5.b}
    \end{subfigure}
    \caption{Attention pattern to images patches in regular image encoder from \bslneC (on the left) and the \tie model (on the right). Brown patches indicate higher attention values and white patches indicate lower values. In both examples above, the \tie model demonstrates stronger focus on the content most relevant to the given question. Examples are from the DocVQA benchmark.}
    \label{fig:fig5}
\end{figure*}

\begin{figure}[ht!]
    \centering
    \begin{subfigure}[t]{0.35\linewidth}
        \centering
        \includegraphics[width=\linewidth]{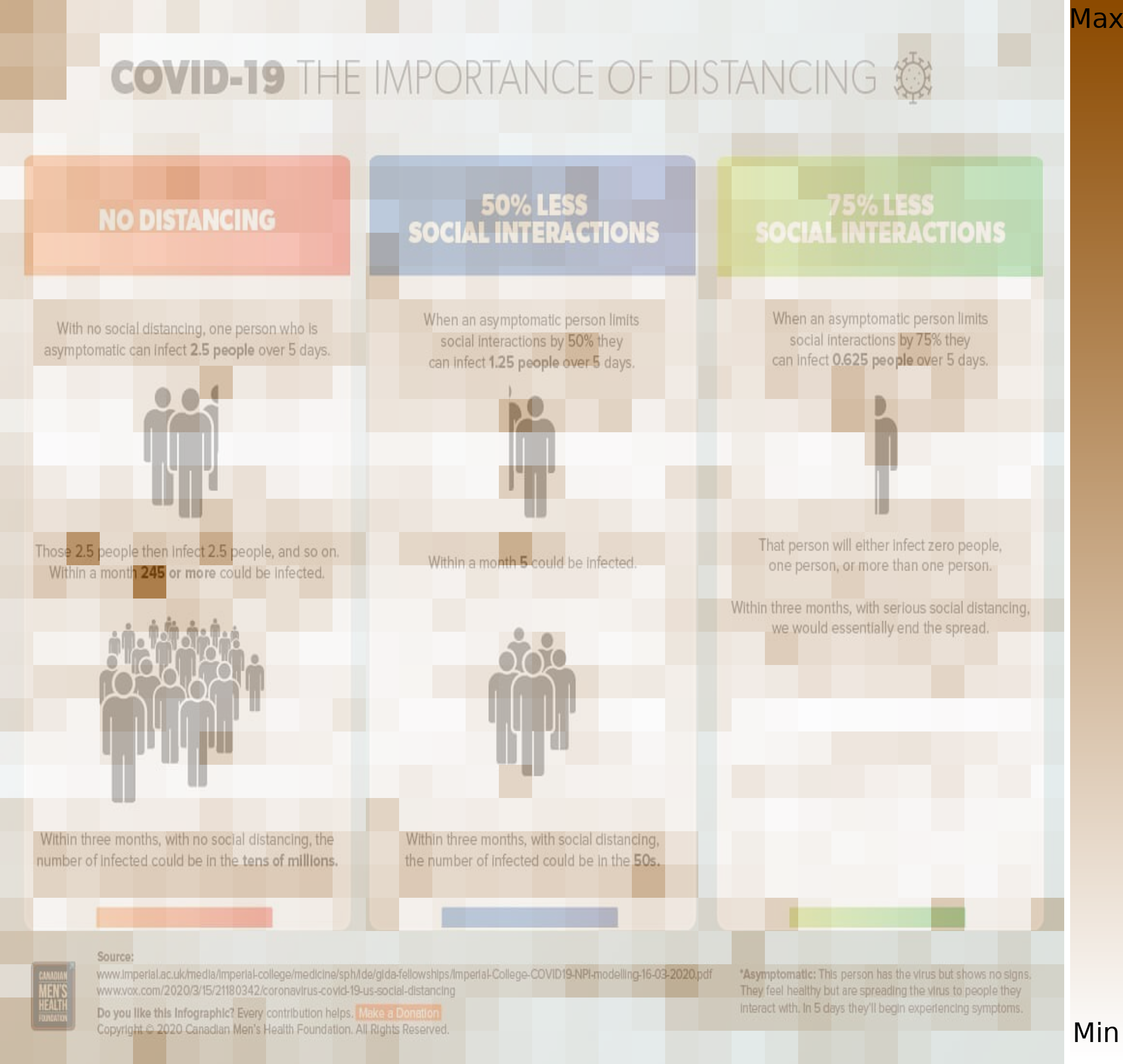}
        \caption{``What is the number of people an asymptotic person infects to if they reduced interaction by 50\%? Provide a short and direct response"}
        \label{fig:fig6.a}
    \end{subfigure}
    \quad\quad\quad
    \begin{subfigure}[t]{0.35\linewidth}
        \centering
        \includegraphics[width=\linewidth]{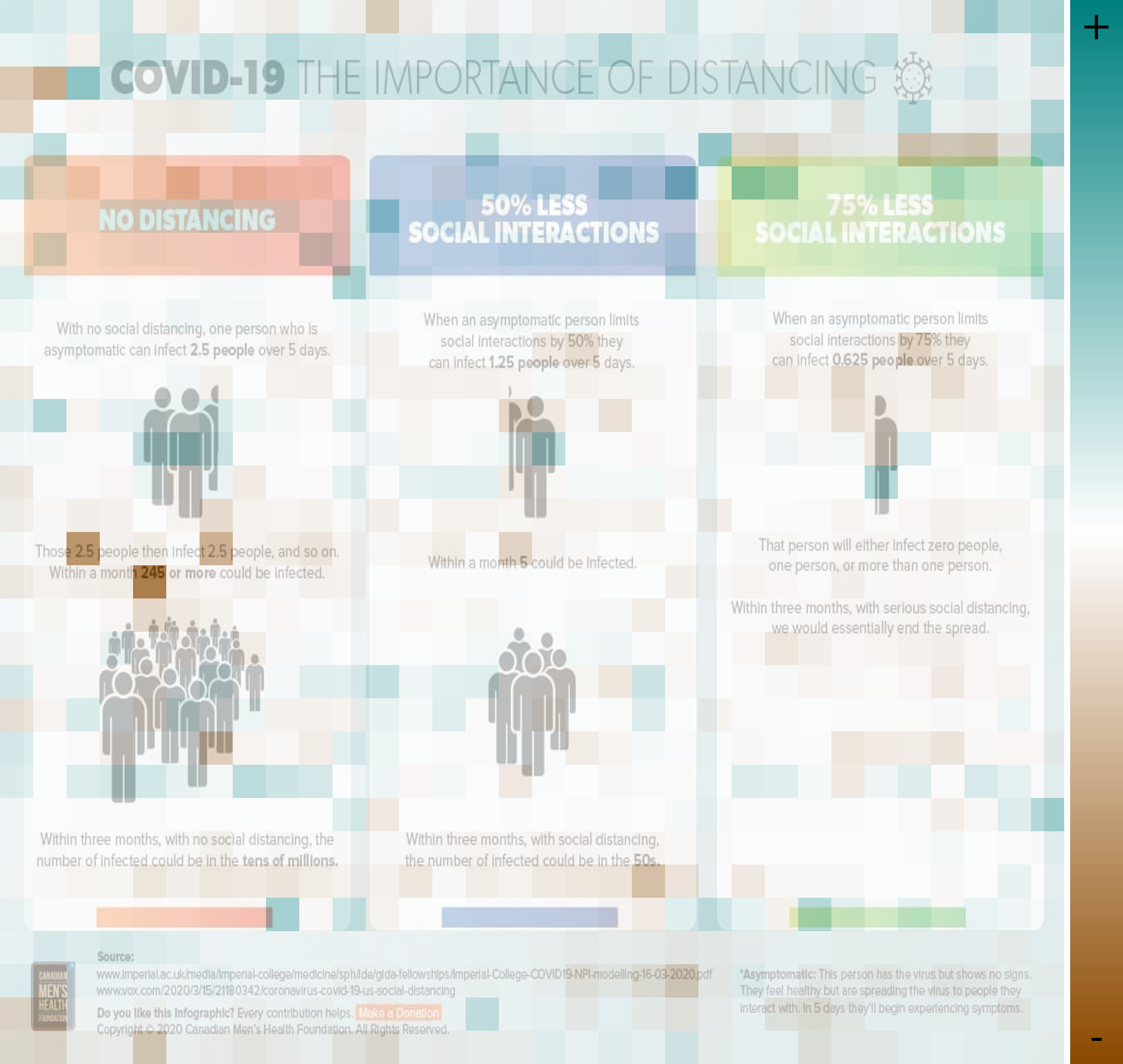}
        \caption{``Summarize the image in a few words. Provide a short and direct response"}
        \label{fig:fig6.b}
    \end{subfigure}
    \caption{(a) When the question pertains to a quantity, the \tie model focusses more on numerical elements within the image (brown=high; white=low) (b) Change in attention (cyan=increase; brown=decrease) when the question is made general. Numeric elements show brown i.e. reduced attention for the generic query}
    \label{fig:fig6}
\end{figure}

\subsubsection{\texorpdfstring{\ours}{PLM(TIE)} is adaptable to Multi-turn Queries}
In some cases, multiple questions may refer to the same image, making repeated re-encoding of that image computationally expensive. In Ablation 2, we test whether a generic query at inference can address this. The generic query is provided only to \tie, while the LLM still receives the original query. We evaluate a trained \ours model under two inference-time prompting configurations. 
\texttt{Generic Query @Inf.}
uses the query ``\textit{Answer the question about the given image: Summarize the details in the image}". 
\texttt{Instruction only @Inf.}
uses the simplified prompt ``\textit{Answer the question about the given image:}" for \tie. While \texttt{Instruction only @ Inf.} does not perform well due to the training-test mismatch of queries, \texttt{Generic Query @ Inf.} maintains a gain of $+1.0$ over \bslneC. This indicates that the text-guided training in \tie more effectively optimizes the image encoder to capture salient visual features, even without explicit query cues, thus making it adaptable to handling multi-turn queries.

\subsubsection{Compute Alone \texorpdfstring{$\neq$}{not equals} Better Image Encoding. Query Conditioning is Crucial.}
Image-text conditioning in \tie adds extra computation compared to standard image encoders. It is important to determine whether the performance gains in \tie arise solely from the additional image–text computation. In Ablation 3, we retrain multiple variants of the model by modifying the text input provided to the \tie architecture. \ours uses both an instruction prompt - ``\textit{Answer the following question about the given image:}’’ and the corresponding question as input to \tie.
\texttt{Question only} version uses only the question, omitting the instruction. \texttt{Instruction only} removes the question while retaining only the instruction.
\texttt{Dummy question} maintains the same number of tokens as the full instruction–question input but replaces the question with a dummy sequence (e.g., “\textit{A A A A ...}”). Note that this variant uses the exact same compute as the default \ours but without query conditioning. The first two methods incorporating query conditioning substantially outperform the latter two without it, indicating that \tie’s improvements stem not from additional computation within the image encoder, but from effective query–image conditioning.

\subsubsection{Better architectures of \tie?}
In Ablation 4, we examine whether alternative configurations of \tie can enhance image–text interaction. We evaluate three configurations: (i) \texttt{T5-Same}, reusing the same \tfl encodings across all ViT layers; (ii) the default \tie, passing them through FFNs for layer-specific adaptation; and (iii) \texttt{Cross Encoder}, adding cross-attention layers between self-attention and FFN blocks for image-query interaction. Empirically, \tie's layer-specific adaptation outperforms the static \texttt{T5-Same} embeddings. Although \texttt{Cross Encoder} achieves further gains over the default \tie, it introduces roughly 30\% more parameters, increasing training time and computational complexity.

\subsection{Qualitative Analysis}
\subsubsection{Improved Attention over the Baseline}

\Cref{fig:fig5} compares the attention patterns of a standard IE from \bslneC with \tie from \ours. Attention patterns were obtained by averaging attention weights across all layers and heads of the image encoder over the image tokens, followed by min–max normalization. In both examples, \tie demonstrates better question-aligned attention i.e. it highlights image regions which contain ``due of "M/s Meenakshi Stores"'' in \Cref{fig:fig5.a} and ``contribution to provident funds'' in \Cref{fig:fig5.b}. This demonstrates \tie{}’s ability to leverage the question to attend selectively to the most informative visual areas.

\subsubsection{Changing Attention with Different Queries}
Here, we examine how the attention distribution varies across different queries applied to the same image.
\Cref{fig:fig6.a} presents the attention map generated by the \tie model when asked a question concerning the number of people. In this case, the model focuses primarily on regions containing numerical information, reflected by the prominent brown highlights.
When the query is broadened to request a summary of the image, the resulting change in attention is shown in \Cref{fig:fig6.b}. The cyan regions indicate increased attention to the textual elements of the image, while the brown regions over the numeric areas reflect a reduction in attention, as expected for a more general question.
This demonstrates \tie’s capacity to dynamically adjust its visual focus in accordance with the semantic intent of the query.

\section{Conclusion}
In this work, we introduced \tie, a query-conditioned image encoder that promotes tighter vision–language integration within the image encoding process. Across multiple architectures, tile setups, and token configurations, \tie consistently outperforms its standard counterparts trained under identical settings. Despite using only half the number of tiles, the proposed \tie model outperforms the standard IE while achieving faster inference. Even when paired with non-contextual text embeddings from the VLM, \tie achieves significant gains, highlighting the strength of this architecture. Simply scaling compute in conventional IEs offers limited gains—text alignment is crucial. \tie also generalizes effectively to generic questions and continues to improve when enhanced with cross-encoder mechanisms or additional compute. Together, the results demonstrate that \tie delivers a substantial step toward question-aware visual modeling, setting a foundation for future work on scalable, jointly optimized multimodal encoders.

\rt{modify the fig 5 attention pattern to show difference.}
\bd{Worth talking about why?}
\bd{PLM itself without finetuning is also a reasonable baseline -- perhaps we can report improvements over that as well?}
\bd{I cannot make any sense of this figure. Why not use a bar plot or line plot?}

{
    \small
    \bibliographystyle{ieeenat_fullname}
    \bibliography{main}
}

\clearpage
\appendix
\crefalias{section}{appendix}
\crefalias{subsection}{appendix}

\section{Replicability}
Source code will be made publicly available.

\section{Limitations}
\paragraph{Larger models not evaluated.} Our analysis is limited to the 1B and 3B models. Training larger models, especially with larger tile configurations was computationally very expensive, exceeding 50 hours per run in some cases, which constrained us to smaller model sizes.

\paragraph{Larger tile configurations not explored.}
We evaluated tile settings up to T=8, but did not experiment with larger configurations such as the T=36 setup used in \plm, primarily due to compute constraints. Although modern VLMs can support up to 36 tiles, in practice—especially for images without extreme aspect ratios—models like \textsc{InternVL3} typically rely on far fewer tiles (around eight) \citep{zhu2025internvl3, chen2024expanding}. This is consistent with our results: the 8-tile performance in \Cref{tab:multi_tiles} narrows the gap to the 36-tile \plm numbers in \Cref{tab:model_sizes}.

\section{Supplementary material}
\subsection{Model, Training and Dataset Details}\label{apsec:resources}
\label{sec:appendix}
\paragraph{Training Details.} We train our image encoder on top of pretrained \pef (\pes) model \citep{bolya2025perception}. \citet{cho2025perceptionlm} used $\pes{-}\rm L$ (0.3B params) as the image encoder and released \plm models of varying scales.
$\pes{-}\rm L$ has a patch size of $P=14$ and a tile size of $H\times W=448\times 448$. The \plm projector applies a spatial downsampling of $D=2$, resulting in 256 tokens per tile. 
In our experiments, the \tfl encoder \citep{raffel2020exploring} (0.3B params) was used as the auxiliary text encoder in \tie. The \tie image encoder with the \plm language model as a whole is referred to as \ours. In each case, the \tie model is trained with a peak learning rate of $1e^{-4}$ with weight decay of $0.01$ for $25K$ steps with an effective batch size of $2048$.
In most cases (except for \Cref{tab:multi_tiles}), both training and inference are conducted with a single tile, i.e., $T{=}1$. Additional details in  \Cref{sec:appendix}. 

\paragraph{Data Details.}We utilize the stage-$2$ training dataset curated by \citet{cho2025perceptionlm}, using only the image samples and excluding any video and text-only data. For simplicity, we retain only the first turn from multi-turn samples and conduct training and evaluation on single-image tasks. Nonetheless, the proposed approach can be easily modified to accommodate multi-turn and interleaved image inputs. The final dataset comprises 37M datapoints each with one image.

\begin{table}[t]
\centering
\resizebox{0.4\textwidth}{!}{%
\begin{tabular}{lrccr}
\toprule
\textbf{Model} & \textbf{T} & \textbf{Enc} & \textbf{Dec} & \textbf{Training Time (hrs)} \\
\midrule
\bslneC & 1 & 0.3B & 1B & 14.5 \\
\bslneC & 1 & 0.3B & 3B & 31.5 \\
\bslneC & 2 & 0.3B & 1B & 28.0 \\
\bslneC & 4 & 0.3B & 1B & 34.0 \\
\bslneC & 8 & 0.3B & 1B & 44.5 \\
\midrule
\ours & 1 & 0.3B & 1B & 17.5 \\
\ours & 1 & 0.3B & 3B & 35.0 \\
\ours & 2 & 0.3B & 1B & 31.5 \\
\ours & 4 & 0.3B & 1B & 44.5 \\
\ours & 8 & 0.3B & 1B & 57.0 \\
\bottomrule
\end{tabular}}
\caption{Training time (in hours) across different encoder–decoder configurations and tile counts $T$.}
\label{tab:resources_runtime}
\end{table}

\begin{table*}[ht!]
\centering
\resizebox{\textwidth}{!}{%
\begin{tabular}{l|c|c|c|rrrrrrrrrrr|r}
\toprule
\textbf{Model} & \textbf{Enc.} & \textbf{Dec.} & $T$ & \textbf{DocVQA} & \textbf{ChartQA} & \textbf{TextVQA} & \textbf{InfoVQA} & \textbf{AI2D} & \textbf{OCRB} & \textbf{MMMU} & \textbf{OKVQA} & \textbf{VIZWIZ} & \textbf{MME} & \textbf{POPE} & \textbf{Avg.} \\
\midrule
\textsc{Qwen2VL-2B}$^{\dagger}$  &$0.67$B&$1.5$B& Dynamic\footnotemark[1]& 90.1 & 75.3 & 80.3 & 65.5 & 84.6 & 80.9 & 41.1 & 59.7 & 67.4 & 69.29 & 87.2 & 72.85 \\
\textsc{InternVL2.5-1B}$^{\dagger}$ &$0.3$B&$0.5$B&up to 36& 84.8 & 75.9 & 72.0 & 56.0 & 77.8 & 78.5 & 40.9 & 51.5 & 47.4 & 69.64 & 90.2 & 67.69 \\
\plm-\textsc{1B}$^{\dagger}$ &$0.3$B&$1$B&up to 36& 90.7 & 78.6 & 82.1 & 63.0 & 84.9 & 80.7 & 34.8 & 61.0 & 59.7 & 57.25 & 88.4 & 71.01 \\\hline
\textsc{Qwen2.5VL-3B}$^{\dagger}$ &$0.67$B&$3$B & Dynamic\footnotemark[1]& 93.9 & 83.1 & 79.3 & 77.1 & 90.2 & 79.7 & 53.1 & 63.2 & 71.9 & 79.32 & 88.2 & 78.09 \\
\textsc{InternVL2.5-4B}$^{\dagger}$ &$0.3$B&$0.5$B&up to 36& 91.6 & 84.0 & 79.3 & 72.1 & 90.5 & 82.8 & 52.3 & 64.0 & 61.8 & 83.46 & 91.0 & 77.53 \\
\plm-\textsc{3B}$^{\dagger}$ &$0.3$B&$3$B&up to 36& 93.8 & 84.3 & 84.3 & 74.6 & 90.9 & 83.0 & 41.2 & 66.8 & 64.0 & 67.11 & 88.4 & 76.22 \\
\midrule\midrule
\plm-\textsc{1B}$^\ddagger$ & \multirow{3}{*}{\shortstack{$0.3$B\\(448/14)}} & \multirow{3}{*}{$1$B} & \multirow{3}{*}{$1$} & 51.05 & 58.56 & 65.34 & 27.25 & 83.32 & 59.20 & 32.67 & 55.76 & \textbf{59.18} & 54.18 & 85.73 & 57.48 \\
\bslneC-\textsc{1B} & & & & 68.61 & 72.28 & 72.57 & 33.92 & \textbf{84.46} & 64.70 & 31.78 & \textbf{56.02} & 57.14 & 58.7 & 87.42 & 62.51 \\
\ours-\textsc{1B} & & & &\textbf{73.92} & \textbf{73.32} & \textbf{72.57} & \textbf{37.32} & 84.36 & \textbf{66.40} & \textbf{32.44} & 55.90 & 57.61 & \textbf{59.11} & \textbf{88.11} & \textbf{63.73} \\
\midrule
\plm-\textsc{3B}$^\ddagger$ & \multirow{3}{*}{\shortstack{$0.3$B\\(448/14)}} & \multirow{3}{*}{$3$B} & \multirow{3}{*}{$1$} &52.83 & 60.48 & 66.42 & 31.84 & 89.05 & 61.00 & 38.78 & 61.51 & 59.20 & 64.01 & 87.61 & 61.16 \\
\bslneC-\textsc{3B} & & & & 72.61 & 76.68 & 73.60 & 38.02 & 89.99 & \textbf{68.60} & 38.44 & 59.37 & 59.19 & 65.93 & 88.33 & 66.43 \\
\ours-\textsc{3B} & & & & \textbf{75.42} & \textbf{76.80} & \textbf{74.28} & \textbf{40.75} & \textbf{90.09} & 66.80 & \textbf{38.67} & \textbf{61.01} & \textbf{63.28} & \textbf{65.99} & \textbf{88.34} & \textbf{67.40} \\
\bottomrule
\end{tabular}%
}
\caption{Evaluation results on all eleven benchmarks. 
For the evaluated models (last six rows), the number of tiles, $T=1$. 
$Enc.$ denotes image encoder’s parameters and $Dec.$ denotes the LLM size.
\ours consistently outperforms both \bslneC and \plm across $1$B, $3$B sizes. $^{\dagger}$-reported in the respective works, \plm \cite{cho2025perceptionlm}. $^{\ddagger}$-\plm trained with 36 tiles (1 tile training during initial stages) evaluated at 1 tile.
}\label{tab:model_sizes_full}
\end{table*}

\begin{table*}[ht!]
\centering
\resizebox{\textwidth}{!}{%
\begin{tabular}{l|c|c|rrrrrrrrrrr|r}
\toprule
\textbf{MultiTile} & \textbf{Dec.} & $T$ & \textbf{DocVQA} & \textbf{ChartQA} & \textbf{TextVQA} & \textbf{InfoVQA} & \textbf{AI2D} & \textbf{OCRB} & \textbf{MMMU} & \textbf{OKVQA} & \textbf{VIZWIZ} & \textbf{MME} & \textbf{POPE} & \textbf{Avg.} \\
\midrule
\plm-\textsc{1B} & \multirow{3}{*}{$1$B} & \multirow{3}{*}{up to $2$} & 54.44 & 61.08 & 66.05 & 36.81 & 84.10 & 62.20 & 32.67 & 55.29 & \textbf{59.14} & 56.16 &	86.3 & 59.48 \\
\bslneC-\textsc{1B} & & & 66.67 & 72.04 & 71.44 & 38.41 & \textbf{85.23} & 65.90 & \textbf{33.78} & 56.56 & 57.08 & 57.47 & 87.61 & 62.93 \\
\ours-\textsc{1B} & & & \textbf{71.78} & \textbf{73.36} & \textbf{71.56} & \textbf{43.96} & 84.88 & \textbf{66.90} & 33.11 & \textbf{56.81} & 57.43 & \textbf{58.16} & \textbf{88.00} & \textbf{64.18} \\
\midrule
\plm-\textsc{1B} & \multirow{3}{*}{$1$B} & \multirow{3}{*}{up to $4$} & 80.92 & 72.44 & 77.59 & 50.16 & 84.46 & 73.50 & 32.11 & 55.83 & \textbf{59.78} & 58.59&	87.09&	66.59 \\
\bslneC-\textsc{1B} & & & 83.70 & 76.92 & \textbf{80.78} & 48.70 & \textbf{85.95} & \textbf{74.40} & 32.44 & 57.43 & 57.80 & 58.30 & 87.29 & 67.61 \\
\ours-\textsc{1B} & & & \textbf{86.72} & \textbf{77.68} & 80.50 & \textbf{57.09} & 85.78 & 73.70 & \textbf{33.67} & \textbf{57.63} & 56.69 & \textbf{60.15} & \textbf{87.87} & \textbf{68.86} \\
\midrule
\plm-\textsc{1B} & \multirow{3}{*}{$1$B} & \multirow{3}{*}{up to $8$} & 81.30 & 74.80 & 79.79 & 54.66 & 85.62 & \textbf{74.70} & 32.44 & 55.88 & \textbf{59.66} & 57.61&	87.38&	67.62 \\
\bslneC-\textsc{1B} & & & 84.11 & 78.44 & \textbf{81.49} & 52.09 & 86.72 & 73.40 & 32.89 & \textbf{57.69} & 57.60 & 58.23 & 87.49 & 68.20 \\
\ours-\textsc{1B} & & & \textbf{86.88} & \textbf{78.56} & 80.76 & \textbf{58.32} & \textbf{86.85} & 73.90 & \textbf{33.11} & 57.13 & 56.64 & \textbf{58.30} & \textbf{87.71} & \textbf{68.92} \\
\bottomrule
\end{tabular}}
\caption{Evaluation results across benchmarks for 2-tile, 4-tile, and 8-tile at 1B scale. \ours outperforms \bslneC and \plm across the settings. Also, \tie($T=4$) beats \bslneC($T=8$). \ours-\textsc{1B} ($T=8$) gets close to the \plm-\textsc{1B} ($T=36$) version from \Cref{tab:model_sizes}.
}\label{tab:multi_tiles_full}
\end{table*}

\paragraph{Models Compared.} \plm models of varying sizes serve as the primary baselines for comparison. \plm\ models were trained with varying tile per image (1 through 16) across different training stages, followed by final tuning with 36 tiles. Trained \plm models (36 tiles) were evaluated at different tile settings during inference. To ensure a fair assessment of the additional compute involved in the training process of \ours (\Cref{ss:training}), we construct a \bslneC variant that continues pretraining the \plm model on identical batches and for the same number of optimization steps as the \ours model. Notably, the image encoder in \ours has the same number of trainable parameters as the corresponding image encoder in \bslneC. Parameters in \tfl are kept frozen as discussed earlier. Note that \ours uses both an instruction prompt - ``\textit{Answer the following question about the given image:}’’ and the corresponding question as input to \tie.

\paragraph{Evaluation Benchmarks.}
We compare \ours with \bslneC and \plm over 14 diverse image-to-text tasks spanning OCR, chart, document, and visual question answering, as evaluated in~\citet{bolya2025perception}. The tasks include – DocVQA \cite{mathew2021docvqa}, ChartQA \cite{zheng2025advancing}, TextVQA \cite{singh2019towards}, InfoVQA \cite{mathew2022infographicvqa}, AI2D \cite{kembhavi2016diagram}, OCRBench \cite{liu2024ocrbench}, MMMU \cite{yue2024mmmu}, OK-VQA \cite{schwenk2022okvqa}, VizWiz \cite{bigham2010vizwiz}, MME \cite{chaoyou2023mme}, POPE \cite{li2023evaluating}, Flickr \cite{young2014image}, COCO \cite{lin2014microsoft}, NoCap \cite{agrawal2019nocaps}. We adopt evaluation protocols consistent with \plm and those followed by LMMS-Eval \cite{zhang2024lmmsevalrealitycheckevaluation, lmms_eval2024}. For OCRBench and MME, scores are further normalized by 1000 (number of datapoints) and 2800 (maximum score), respectively, to bring them to a similar scale (out of 100) for averaging with other datasets.

\subsection{Training Resources and Runtime Analysis}
All experiments were performed on $64$ H$200$ gpus.
\Cref{tab:resources_runtime} reports the training time across different configurations of \bslneC\ and \ours.

\subsection{Additional VQA Datasets}\label{ap:additional_datasets}
\Cref{tab:model_sizes_full,tab:multi_tiles_full} extend \Cref{tab:model_sizes,tab:multi_tiles} by including three additional visual question answering datasets datasets — AI2D, MMMU, and POPE. On these datasets, the performance of \bslneC and \ours remains largely comparable, with \ours showing a modest gain in most cases. 

\subsection{Additional Captioning Datasets}\label{appsub:captioning_data}
\Cref{tab:coco_nocaps_flickr} shows the captioning performance of \bslneC and \ours. Captioning datasets have same question across all examples. \ours\ maintains comparable performance to \bslneC (worse on FLICKR and better on NOCAPS).

\begin{table*}[ht!]
\centering
\resizebox{\textwidth}{!}{%
\begin{tabular}{ll|rrrrrrrrrrr|r}
\toprule
\textbf{Abl.} & \textbf{Model} & \textbf{DocVQA} & \textbf{ChartQA} & \textbf{TextVQA} & \textbf{InfoVQA} & \textbf{AI2D} & \textbf{OCRB} & \textbf{MMMU} & \textbf{OKVQA} & \textbf{VIZWIZ} & \textbf{MME} & \textbf{POPE} & \textbf{Avg.} \\\midrule
\multirow{3}{*}{1} &
\bslneC-\textsc{1B} & 68.61 & 72.28 & 72.57 & 33.92 & 84.46 & 64.70 & 31.78 & 56.02 & 57.14 & 58.7&	87.42&	62.51 \\
& \ours-\textsc{1B} (\texttt{with VLM's TE}) & 72.43 & 72.80 & 72.71 & 36.57 & 84.97 & 65.30 & 33.00 & 56.62 & 55.30 & 59.65&	88.14&	\underline{63.41} \\
& \ours-\textsc{1B} (with \tfl) & 73.92 & 73.32 & 72.57 & 37.32 & 84.36 & 66.40 & 32.44 & 55.90 & 57.61 & 59.11&	88.11&	\textbf{63.73} \\
\midrule
\multirow{3}{*}{2} & \ours-\textsc{1B} & 73.92 & 73.32 & 72.57 & 37.32 & 84.36 & 66.40 & 32.44 & 55.90 & 57.61 & 59.11&	88.11&	\textbf{63.73} \\
& \ours-\textsc{1B} (\texttt{Generic Question @Inference}) & 71.33 & 73.32 & 71.99 & 35.81 & 84.46 & 65.80 & 32.67 & 55.97 & 58.69 & 59.39&	88.1&	\underline{63.41} \\
& \ours-\textsc{1B} (\texttt{Instruction only @Inference}) & 68.92 & 72.88 & 70.41 & 33.80 & 84.59 & 64.70 & 32.44 & 55.94 & 58.25 & 58.94&	87.12&	62.54
  \\\midrule
\multirow{4}{*}{3}  & \ours-\textsc{1B} & 73.92 & 73.32 & 72.57 & 37.32 & 84.36 & 66.40 & 32.44 & 55.90 & 57.61 & 59.11&	88.11&	\textbf{63.73} \\
 & \ours-\textsc{1B} (\texttt{Question only})& 73.40 & 72.60 & 73.03 & 36.91 & 84.33 & 66.30 & 33.33 & 56.85 & 57.52 & 59.84&	87.61&	\underline{63.79} \\
& \ours-\textsc{1B} (\texttt{Instruction only}) & 68.90 & 72.48 & 72.46 & 33.56 & 84.16 & 64.80 & 32.44 & 56.72 & 57.53 & 57.13&	87.56&	62.52 \\
& \ours-\textsc{1B} (\texttt{Dummy Question}) & 67.32 & 71.84 & 72.18 & 34.56 & 83.84 & 66.30 & 33.11 & 56.40 & 55.75 & 57.93&	87.21&	62.40 \\\midrule 
\multirow{3}{*}{4} & \ours-\textsc{1B} (\texttt{T5 Same}) & 73.69 & 72.68 & 73.05 & 37.90 & 84.84 & 65.80 & 32.00 & 56.32 & 57.44 & 58.07&	87.66&	63.59 \\
& \ours-\textsc{1B} & 73.92 & 73.32 & 72.57 & 37.32 & 84.36 & 66.40 & 32.44 & 55.90 & 57.61 & 59.11&	88.11&	\underline{63.73} \\
& \ours-\textsc{1B} (\texttt{Cross Encoder}) & 74.25 & 73.12 & 72.94 & 37.31 & 84.62 & 69.10 & 32.44 & 56.58 & 57.02 & 58.57&	87.34&	\textbf{63.94} \\\midrule
\multirow{2}{*}{5} 
& \bslneC-\textsc{1B} (\texttt{50k steps})&
71.24 & 73.00 & 73.05 & 34.12 & 84.46 & 66.90 & 32.22 & 56.10 & 57.07 & 58.32&	87.42&	63.08 \\
& \ours-\textsc{1B} (\texttt{50k steps})&
75.93 & 74.60 & 73.31 & 38.40 & 84.49 & 67.30 & 32.11 & 56.20 & 57.00 & 58.59&	87.8&	64.16 \\\midrule
\multirow{2}{*}{6} & \bslneC-\textsc{1B} (\texttt{on Cambrian}) & 66.72 & 72.64 & 63.53 & 33.12 & 79.08 & 60.90 & 31.56 & 51.52 & 62.89 & 60.95 & 72.97 & 59.63 \\
& \ours-\textsc{1B} (\texttt{on Cambrian}) & 66.62 & 73.16 & 67.16 & 33.51 & 84.81 & 63.10 & 31.56 & 54.78 & 53.78 & 64.20 & 86.84 & \textbf{61.77} \\
\bottomrule
\end{tabular}}
\caption{Ablation studies - 1. Using VLM's non-contextual embeddings already results in notable boost over the baseline. 2. \ours when used with a generic question at test time performs better than \bslneC. 3. Merely increasing compute inside the image encoder is not sufficient. Question conditioning is important. 4. Adding more parameters to the \tie image encoder might be beneficial. 5. \ours beats \bslneC across different training setups and 6. across different training distributions. }\label{tab:ablations_full}
\end{table*}

\begin{table*}[htbp]
\centering
\resizebox{0.4\textwidth}{!}{%
\begin{tabular}{lccc}
\toprule
\textbf{Model} & \textbf{COCO} & \textbf{NOCAPS} & \textbf{FLICKR30K} \\
\midrule
\bslneC-\textsc{1B} & 136.08 & 124.60 & \textbf{74.48} \\
\ours-\textsc{1B} & \textbf{136.09} & \textbf{125.94} & 71.67 \\\hline
\bslneC-\textsc{3B} & \textbf{137.46} & 123.83 & \textbf{80.92} \\
\ours-\textsc{3B} & 136.72 & \textbf{124.64} & 78.24 \\
\bottomrule
\end{tabular}}
\caption{Performance comparison across COCO, NOCAPS, and FLICKR30K datasets.}
\label{tab:coco_nocaps_flickr}
\end{table*}

\begin{table*}[ht!]
\centering
\resizebox{\textwidth}{!}{%
\begin{tabular}{l|rrrrrrrrrrr|r}
\toprule
\textbf{Model} & \textbf{DocVQA} & \textbf{ChartQA} & \textbf{TextVQA} & \textbf{InfoVQA} & \textbf{AI2D} & \textbf{OCRBench} & \textbf{MMMU} & \textbf{OKVQA} & \textbf{VIZWIZ} & \textbf{MME} & \textbf{POPE} & \textbf{Avg} \\
\midrule
\bslneC (L=192, N=64) & 47.82 & 61.36 & 64.31 & 26.93 & 80.54 & 54.00 & 32.89 & 54.74 & 55.69 & 56.69 & 86.94 & 56.54 \\
\ours (L=192, N=64) & 57.46 & 63.08 & 64.52 & 27.35 & 79.57 & 55.10 & 32.78 & 54.36 & 53.70 & 54.83 & 87.19 & 57.27 \\\hline
\bslneC (D=4, N=64) & 50.70 & 62.88 & 66.70 & 27.60 & 80.28 & 56.30 & 32.44 & 54.49 & 55.33 & 55.81 & 86.34 & 57.17 \\
\ours (D=4, N=64) & 60.95 & 64.36 & 67.64 & 29.72 & 79.99 & 57.60 & 32.89 & 54.78 & 55.21 & 56.03 & 86.38 & 58.69 \\\hline
\bslneC (L=128, N=128) & 59.09 & 67.64 & 68.18 & 30.47 & 82.77 & 58.40 & 32.67 & 55.51 & 56.09 & 56.79 & 87.10 & 59.52 \\
\ours (L=128, N=128) & 66.06 & 68.28 & 68.64 & 31.61 & 81.54 & 61.00 & 32.67 & 54.95 & 54.63 & 57.61 & 87.36 & 60.40 \\
\bottomrule
\end{tabular}}
\caption{Under settings that use fewer tokens per image, \ours\ performs better across all approaches.}
\label{tab:fewer_tokens_full}
\end{table*}

\begin{table*}[ht!]
\centering
\resizebox{\textwidth}{!}{%
\begin{tabular}{l|rrrrrrrrrrr|r}
\toprule
\textbf{Model} & \textbf{DocVQA} & \textbf{ChartQA} & \textbf{TextVQA} & \textbf{InfoVQA} & \textbf{AI2D} & \textbf{OCRBench} & \textbf{MMMU} & \textbf{OKVQA} & \textbf{VIZWIZ} & \textbf{MME} & \textbf{POPE} & \textbf{Avg.} \\
\midrule
\bslneC (L=64) & 44.90 & 60.40 & 64.30 & 25.75 & 79.95 & 49.70 & 30.67 & 54.38 & 56.08 & 55.78 & 86.11 & 55.27 \\
\ours (L=64) & 53.04 & 61.68 & 65.73 & 27.36 & 78.21 & 53.70 & 32.67 & 54.79 & 55.21 & 57.32 & 86.42 & 56.92 \\\hline
\bslneC (L=128) & 57.71 & 67.40 & 67.29 & 30.74 & 82.09 & 56.90 & 31.22 & 55.07 & 56.33 & 57.24 & 87.11 & 59.01 \\
\ours (L=128) & 64.49 & 67.36 & 68.81 & 31.52 & 80.99 & 60.10 & 32.78 & 55.46 & 54.88 & 57.88 & 87.61 & 60.17 \\\hline
\bslneC (L=192) & 62.22 & 68.24 & 68.18 & 31.75 & 83.16 & 60.60 & 32.11 & 55.34 & 56.20 & 57.58 & 87.93 & 60.30 \\
\ours (L=192) & 68.08 & 68.88 & 69.43 & 33.10 & 82.19 & 63.60 & 32.33 & 55.30 & 54.84 & 57.28 & 87.93 & 61.18 \\\hline
\bslneC (L=256) & 64.30 & 68.96 & 68.73 & 32.52 & 83.39 & 61.00 & 32.56 & 55.39 & 55.98 & 56.98 & 87.97 & 60.71 \\
\ours (L=256) & 69.77 & 69.96 & 69.53 & 33.31 & 82.29 & 63.60 & 32.44 & 55.56 & 54.62 & 57.11 & 88.02 & 61.47 \\
\bottomrule
\end{tabular}}
\caption{Train time $L$ randomly selected in $[0,64,128,192]$ for \bslneC and \ours. Models are then tested at different $L$ values at inference.}
\label{tab:matryoshka_full}
\end{table*}

\subsection{Fewer tokens per Image}
\Cref{tab:fewer_tokens_full,tab:fewer_tokens_full} show dataset wise results for \Cref{ss:fewer_tokens}.

\subsection{Ablations on Additional Datasets}
Similar to \Cref{ap:additional_datasets},
\Cref{tab:ablations_full} adds more datasets to those in \Cref{tab:ablations}. We also include the following ablations to further support our claim.

\subsubsection{Continuing Training}
In Ablation 5 of \Cref{tab:ablations_full}, we analyze whether extending training can further narrow the gap between \bslneC and \ours. While all main experiments were trained for 25k steps, we double the training duration to 50k steps. As shown in Ablation~4, \ours continues to maintain a substantial advantage over \bslneC, achieving an average gain of +1.06 compared to +1.2 in the 25k-step setting.

\subsubsection{Varying the Training Data Distribution} 
Given that our work introduces an architectural modification, it is essential to verify its effectiveness across different training data distributions. In Ablation 7, we replace the original 37M \plm dataset with the 5M \textit{Cambrian} dataset from \citet{tong2024cambrian}. Despite this shift in data distribution, \ours continues to significantly outperform \bslneC, demonstrating the robustness of the \tie architecture across diverse training data.

\end{document}